\documentclass{article}

\usepackage[preprint]{neurips_2026}

\usepackage[utf8]{inputenc}
\usepackage[T1]{fontenc}
\usepackage{hyperref}
\usepackage{url}
\usepackage{booktabs}
\usepackage{amsfonts}
\usepackage{amsmath}
\usepackage{nicefrac}
\usepackage{microtype}
\usepackage{graphicx}
\usepackage{xcolor}

\title{Change the Funnel, Not Who May Submit:\\Engineering Peer Review for the Flood}

\author{%
  Mikhail L. Arbuzov\\
  Independent Researcher\\
  \texttt{mike.arbuzov54@gmail.com}
  \And
  Lee Mosbacker\\
  Independent Researcher\\
  \texttt{lee.mosbacker@gmail.com}
  \And
  Sisong Bei\\
  Independent Researcher\\
  \texttt{qurining@gmail.com}
}

\usepackage{float}
\makeatletter
\g@addto@macro\appendix{%
\begin{figure}[H]
\centering
\includegraphics[width=\textwidth]{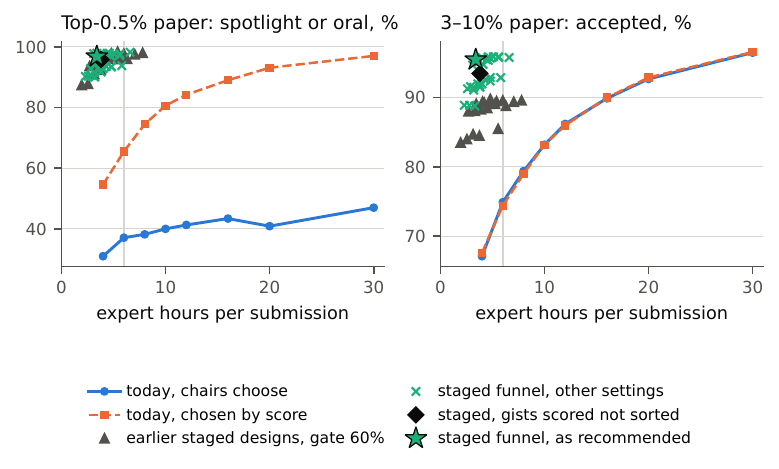}
\caption{Every design simulated: expert hours per submission against placement (computed). Left: a top-0.5\% paper's chance of becoming a spotlight or oral. Right: a 3--10\% paper's chance of acceptance. Today's funnel is plotted with more full reviews per paper, its spotlights and orals chosen by the chairs or by the reviewers' mean scores; the earlier staged designs pass 60\% at the gate and score their gists (Table~\ref{tab:frontier}). Some of the staged funnel's other settings plot higher than the recommended star, which is recommended for its margin under noisier readers (Section~\ref{sec:design}; Appendix~\ref{app:design-search}). The vertical line marks today's 6 hours. Today's 3--10\% acceptance there plots at about 75, against 74 in the text: the two come from different runs, within simulation error. The staged funnel's figures rest on an assumed noise for its new kinds of judgment (Sections~\ref{sec:model} and~\ref{sec:risks}).}
\label{fig:frontier}
\end{figure}
\begin{table}[H]
\caption{The numbers behind Figure~\ref{fig:placement} (computed): percent of each quality tier reaching each outcome or better, today and under the staged funnel, and the reproducibility of each set. The staged columns come from Tables~\ref{tab:compare} and~\ref{tab:tiers} in Appendix~\ref{app:results}.}
\label{tab:buys}
\centering
\small
\setlength{\tabcolsep}{4.5pt}
\begin{tabular}{lrrrrrr}
\toprule
 & \multicolumn{3}{c}{Today's funnel, 6.0 hours} & \multicolumn{3}{c}{Staged funnel, 3.4 hours} \\
\cmidrule(lr){2-4} \cmidrule(lr){5-7}
Quality tier & Accepted & Spotlight & Oral & Accepted & Spotlight & Oral \\
 & & or oral & & & or oral & \\
\midrule
Top 0.5\%   & 98 & 37  & 8   & 99.7 & 97  & 46  \\
0.5--3\%    & 90 & 21  & 3   & 99   & 63  & 5   \\
3--10\%     & 74 & 12  & 1.4 & 95.5 & 15  & 0.1 \\
10--25\%    & 51 & 6   & 0.6 & 68.5 & 0.8 & 0   \\
25--50\%    & 26 & 2   & 0.2 & 18   & 0   & 0   \\
Bottom half & 5  & 0.4 & 0   & 0.3  & 0   & 0   \\
\midrule
Reproducibility & 0.505 & 0.114 & 0.031 & 0.726 & 0.544 & 0.390 \\
\bottomrule
\end{tabular}
\end{table}}
\makeatother

\begin{document}

\maketitle

\begin{abstract}
Every reviewer knows the feeling: more papers each cycle, three full reviews for each of them, and a suspicion that near the accept line the outcome is luck. The numbers bear it out. NeurIPS received 30,709 submissions in 2026, more than three times as many as in 2021, and reviewing them took some 184,000 expert hours, or 92 years of researchers' time. Three-quarters of those hours went to papers that were rejected. And when NeurIPS twice had the same papers reviewed by two committees, the second rejected half of what the first accepted.

Conferences are responding by limiting who may submit. We simulated that answer, and an alternative, in a model calibrated to the NeurIPS experiments. The limits save little and select no better, and those based on a past acceptance fall mostly on new authors. The alternative changes how papers are judged, spending full reads only where a decision turns on them. In a staged funnel, an automated quality check first turns back papers that are not ready. The rest are ranked by their value, from short summaries, by readers that each submission brings. Only the finalists get a full read, which checks their claims and settles the final ranking. The staged funnel needs a little over half of today's expert hours, accepts 95.5\% of papers in the top 3--10\% (against 74\% today), and makes nearly every top-0.5\% paper a spotlight or oral. And because every paper brings its own reader, the funnel keeps pace with the flood.
\end{abstract}

\section{Introduction}
\label{sec:intro}

Submissions to the largest machine-learning conferences have become a flood. NeurIPS received 9,122 submissions in 2021, 21,575 in 2025 and 30,709 in 2026, 42\% more in the last year alone \citep{beygelzimer2023,neurips2025stats,riken2026}. Submissions to ICLR 2027 look set to exceed 60,000 \citep{nature2026caps}. \emph{Science} ties the surge to AI, and \emph{Nature} describes the conferences as inundated \citep{science2026caps,nature2026caps}; anyone who reviews for them has watched more papers arrive each cycle for the same reviewers. The conferences are answering by limiting who may submit. For 2027, ICLR limits authors who have never published at a major venue, because qualified reviewers are scarce, and caps submissions at 20 per author, to curb low-quality papers and to stop co-authorship being used to get around that limit \citep{iclr2027}; CVPR caps submissions at 25 per author \citep{cvpr2027}; and both journals report caps spreading across AI conferences \citep{nature2026caps,science2026caps}. The limits treat the flood as a problem of how many papers arrive.

It is a problem of how review is built. In today's funnel, about three reviewers read every paper in full, and their scores, with an area chair's judgment, decide its fate. At our working figure of two hours for a review done properly, NeurIPS 2026 took about 92,000 full reviews and 184,000 expert hours, or 92 researcher-years at 2,000 hours a year. The reviewers are researchers, so those hours come out of research, and every new submission adds six more. Conferences do ask each submission to bring a reviewer from among its authors \citep{iclr2027,neurips2025rri,acl2026}. But ICLR and ACL accept only authors with a qualifying record, and one such author can cover more than one submission. A paper may have several authors and still bring no one: by ICLR's own count, about a fifth of its 2026 submissions came without a reciprocal reviewer \citep{iclr2027}. Qualified reviewers are scarce, as ICLR itself says \citep{iclr2027}: the reviewing falls on the same few, and conferences recruit reviewers beyond them to cover the rest. So the hours grow with the papers while reviewers lag.

Even before the flood, those hours did not buy a sound selection. NeurIPS has twice had a sample of its submissions reviewed by two independent committees, in 2014 and in 2021, and both times the second committee rejected about half of the papers the first had accepted \citep{cortes2021,beygelzimer2023}. Near the accept line, whether a paper got in depended largely on which reviewers it drew. At the top the committees agreed less still: in 2021, on 3 of the 25 to 29 papers each chose for a spotlight or an oral \citep{beygelzimer2023}. A limit on who may submit leaves all of this in place. Every paper it admits still gets three noisy reviews, from reviewers who do not grow in number with the papers. And a rule that rations entry by a past acceptance falls on new authors, who have none yet.

To change the funnel instead, a conference needs to know what a change would do before adopting it. That is the practice of \emph{decision-making engineering}: simulating the outcomes of a decision process to choose how to change it. We apply it to peer review. For each design we simulate its expert hours and its \emph{placement}: how closely the papers it accepts and highlights follow readers' shared judgment of their quality. Our model is calibrated to the two NeurIPS experiments, which measured how far readers agree, not whether they are right (Section~\ref{sec:model}).

\begin{figure}[t]
\centering
\includegraphics[width=\textwidth]{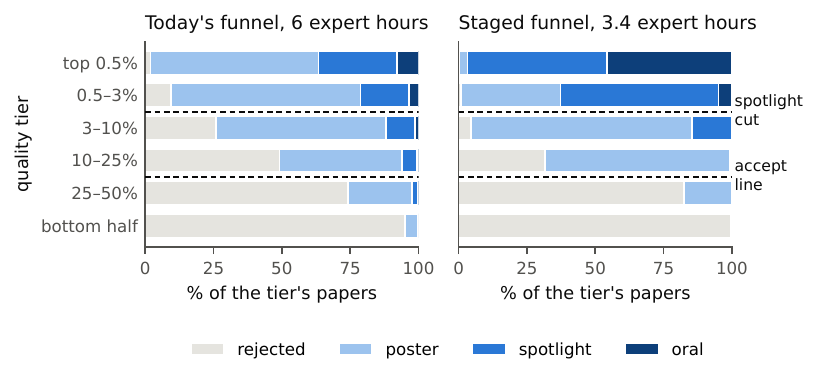}
\caption{Where the papers of each quality tier (Section~\ref{sec:model}) end up (computed). Accepted papers are posters, spotlights or orals. The staged funnel's one weak spot is the 3--10\% row, just below the spotlight cut, where it promotes slightly more papers than today's funnel (Section~\ref{sec:buys}). Its figures rest on an assumed noise for its new kinds of judgment (Sections~\ref{sec:model} and~\ref{sec:risks}). Numbers: Table~\ref{tab:buys}, page~\pageref{tab:buys}.}
\label{fig:placement}
\end{figure}

Simulated this way, today's funnel misplaces papers at the accept line and at the top, and more reviews of the same kind could fix that only at a price no conference would pay (Section~\ref{sec:today}). The limits change who enters the funnel but not how it judges, so they relieve little of the load and select no better, and most of the papers the past-acceptance rules block are new authors' (Section~\ref{sec:limits}). A funnel built differently does better. In the \emph{staged funnel} we work through as a case (Section~\ref{sec:design}), an optional AI gate first turns back papers not ready for review. Each submission names one of its authors as a \emph{peer} to read for it, so the peers grow in number with the submissions. Peers rank short summaries of the papers, \emph{gists}, by sorting them, and only the finalists receive full reviews. Each part has precedent, among them proposers who rank one another at observatories \citep{merrifield2009,alma2026}, summary-first review at ASPLOS and AAAI \citep{kozyrakis2021,leytonbrown2024}, a reviewer required for each submission at ICLR, NeurIPS and ACL \citep{iclr2025,neurips2025rri,acl2026}, and AI review at AAAI-26 and Agents4Science \citep{biswas2026,bianchi2025}; Appendix~\ref{app:related} traces them and states what is new. For a little over half of today's expert hours, the staged funnel places papers better than today's funnel or any limit, and more reproducibly than today's, and it leaves a conference free to choose how selective to be (Figure~\ref{fig:placement}; Section~\ref{sec:buys}). Its lead survives large errors in what it assumes (Section~\ref{sec:risks}). Its price falls on authors: a paper cut before the finalists receives no written review. Conferences should change how their funnel judges papers, not who may submit, and simulate a change before adopting it.

\section{What the numbers rest on}
\label{sec:model}

Every figure in this paper comes from one model of how readers judge papers, to which Section~\ref{sec:limits} adds authors. A \emph{reader} is anyone who judges a paper: a reviewer in today's funnel, and a peer or the AI gate in the staged funnel. In the model, placing papers better means landing them closer to where readers' shared judgment of their quality puts them. The model is calibrated to today's funnel; for the staged funnel it assumes the noise of the new kinds of judgment, which Section~\ref{sec:risks} tests.

In the model, a paper's \emph{quality} is what all readers' judgments of it share. We group papers into six \emph{quality tiers} by rank among all submissions: the top 0.5\%, 0.5--3\%, 3--10\%, 10--25\%, 25--50\% and the bottom half. Three cuts follow NeurIPS 2025's rates \citep{neurips2025stats,riken2025}: the accept line at 24.5\% of submissions, the spotlight cut at 3.2\%, NeurIPS's rate of spotlights, above which a paper is a spotlight or an oral, and the oral cut at 0.36\%. A design's placement is how often each tier's papers land on the side of each cut where they belong, and we report it by tier: how often a 3--10\% paper is accepted, for instance.

Placement asks where papers land; \emph{reproducibility} asks whether they would land there again. It is the share of a design's accepted list that a rerun on the same papers, with fresh readers, accepts again. Two committees that each accepted a quarter of the papers at random would share about a quarter of their lists. We report reproducibility beside placement rather than as part of it, because a funnel can repeat its choices more often and still place papers worse (Section~\ref{sec:risks}).

The model follows Decision Circuits \citep{circuits2026} in treating review as gates fed by readers. Each paper's quality is drawn from a standard normal distribution, and every judgment of the paper, whether a full review, a place in a gist sort or the AI gate's verdict, sees that quality plus fresh noise. We state noise as a variance, in units of quality's, and fit two levels of it to the NeurIPS experiments. The first is the reviews' own. In 2014 the two committees' mean calibrated scores for 166 duplicated papers correlated at 0.55 \citep{cortes2021}, and a review noise of 2.45 makes the means of three full reviews correlate at that value. The second is the chairs', added when they choose spotlights and orals: a variance of 2.0 reproduces the 2021 committees' agreement on those papers, about 0.11. Fitted this way, the model matches two results it was not fitted to. Its accepted list reproduces at 0.505; the committees' lists overlapped at 0.50 in 2014 and 0.49 in 2021, over the 882 duplicated papers still under review \citep{cortes2021,beygelzimer2023}. And spotlights and orals chosen by the reviewers' mean scores alone reproduce at 0.24, against about 0.25 for that choice in 2021 \citep[read from Fig.~3]{beygelzimer2023}. Numbers with a citation are measured; placement and reproducibility are computed, with their simulation error in Appendix~\ref{app:model}.

For the staged funnel, three kinds of judgment have no measurement behind them: how well a peer places a gist within a sort, whether peers drawn from the submissions read as well as today's reviewers, and how well the AI gate judges a paper. The main runs give each the noise of one full review today. And because the experiments measured only how far readers agree, a taste that every reader shares counts in the model as quality, whether or not it is right.

\section{Where a paper lands today}
\label{sec:today}

Today's funnel spends the same three noisy full reviews on every paper. That is why it misplaces papers at the accept line and, once the chairs add their own noise, at the top, and why buying more of the same costs far more than it returns.

Near the accept line, papers differ little in quality, and three noisy reviews cannot tell them apart, so the funnel errs in both directions (Figure~\ref{fig:placement}, left). It turns away papers that belong above the line: a 3--10\% paper is accepted 74\% of the time, and a 10--25\% paper 51\%. In their place it admits papers from below: a bottom-half paper is accepted 5\% of the time, so about one accepted paper in ten comes from the bottom half. A rerun with fresh reviewers accepts only half of the list again (0.505), as the second NeurIPS committee did. To compare many designs at once, we sum this up as the \emph{hit rate at the line}: the share of the papers truly in the top quarter that are accepted. Because the number accepted is fixed, it counts both kinds of error, since each paper from above the line that a design misses lets in one from below. Today's funnel scores 0.63.

At the top, the chairs' noise compounds the reviews'. A top-0.5\% paper is almost always accepted (98\%), but the chairs make it a spotlight or an oral only 37\% of the time, and a 0.5--3\% paper 21\% of the time. A rerun repeats about a tenth of the spotlights and orals (0.11). The second summary, the \emph{hit rate at the top}, is the share of the papers truly in the top 2.5\% that become spotlights or orals, and today's funnel scores 0.20. Both summaries come from the design search of Appendix~\ref{app:design}, which rounds the cuts to 25\% and 2.5\%; the tiers keep NeurIPS's rates.

\begin{figure}[t]
\centering
\includegraphics[width=\textwidth]{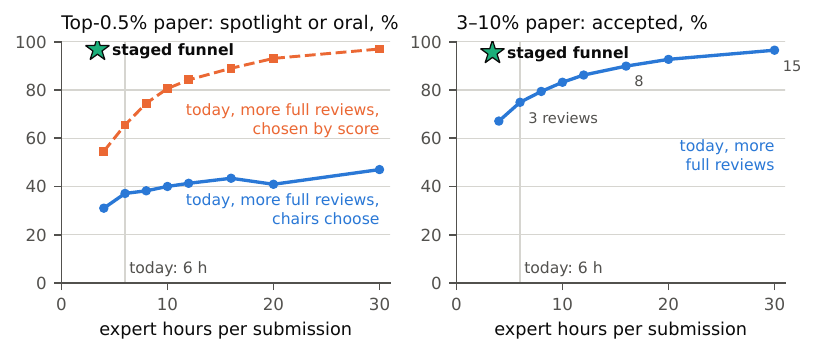}
\caption{What placement costs (computed). Today's funnel buys placement with more full reviews per paper, labelled by number, and at the top by choosing spotlights and orals by the reviewers' mean scores, not by the chairs. Today's 3--10\% acceptance at 6 hours plots at about 75, not the text's 74: a different run, within simulation error. The dip at 20 hours is simulation noise. Table~\ref{tab:frontier} in Appendix~\ref{app:results} gives the points; Figure~\ref{fig:frontier} (page~\pageref{fig:frontier}) adds every design simulated.}
\label{fig:cost}
\end{figure}

Spending alike also puts most of the hours on papers the funnel will reject, since it reviews every paper the same way and accepts about a quarter of them. So each review added to every paper costs two more hours per submission, spent mostly on papers whose fate it will not change, and it buys placement slowly (Figure~\ref{fig:cost}). Eight full reviews per paper, 16 hours or nearly three times today's, raise a 3--10\% paper's acceptance from 74\% to 90\%, but a top-0.5\% paper's chance of a spotlight or oral only from 37\% to 43\%, because no number of reviews removes the chairs' noise. A remedy has to change how the funnel spends its judgment, and admitting fewer papers leaves that as it is.

\section{What the limits do}
\label{sec:limits}

Conferences have adopted two kinds of limit: caps on how many papers an author may submit, and rules that ration entry by a past acceptance. Both change who enters the funnel, not how it judges. So they save little of the load and place papers no better; bounds stricter than any adopted help only at a cost to the best papers; and the rules that ration by a past acceptance fall mostly on new authors.

We simulate each kind of limit beside a stricter bound of our own, to see how far a limit must go before it matters. The adopted caps bind almost no one: by ICLR's own count, fewer than 0.2\% of its authors submitted more than 20 papers in the previous cycle \citep{iclr2027}. So we simulate a far stricter cap, the \emph{self-limit of two}, under which each lead author submits only the two papers they rate highest. For past acceptance we simulate \emph{ICLR's new-author rule} for 2027, under which an author who has never published at a major venue may submit only one paper on which no author is a qualified reciprocal reviewer \citep{iclr2027}; in the model, a past acceptance stands in for both conditions. Beside this rule, the \emph{strict rule} admits a paper only if an author has a past acceptance. For these runs the model adds authors whose ability predicts their papers' quality, and runs today's funnel for three cycles so some earn a past acceptance (Appendix~\ref{app:authors}). Because a limit keeps some papers from being submitted, every figure here is counted against all papers written, and the best papers are the top 3\%, the top two quality tiers. Table~\ref{tab:limits} uses a correlation of 0.5 between ability and quality; 0.3 and 0.7 change no conclusion (Table~\ref{tab:authors} in Appendix~\ref{app:authors}).

\begin{table}[ht]
\caption{Limits on who may submit, under today's funnel, in the model with authors (computed at its third cycle, correlation 0.5, six runs), against all papers written. ``Top 3\%'': the top two quality tiers. ``Newest authors'': the share of blocked papers led by authors who entered in the latest cycle.}
\label{tab:limits}
\centering
\footnotesize
\setlength{\tabcolsep}{2.5pt}
\begin{tabular}{lrrrrrrrr}
\toprule
Limit & Sub- & Hours per & Top 3\%: & Top 3\%: & Top 0.5\%: & Bottom half's & Mean quality: & Newest \\
 & mitted & paper & submitted & accepted & spotlight & share of & blocked vs & authors \\
 & & written & & & or oral & accepted & submitted & \\
\midrule
No limit & 100\% & 6.0 & 100\% & 91\% & 37\% & 10.4\% & -- & -- \\
Strict rule & 90.2\% & 5.4 & 90\% & 84\% & 36\% & 11.3\% & $-0.01$ vs $0.00$ & 68\% \\
ICLR's new-author rule & 96.4\% & 5.8 & 97\% & 89\% & 38\% & 10.7\% & $0.00$ vs $0.00$ & 76\% \\
Self-limit of two & 75.6\% & 4.5 & 94\% & 89\% & 34\% & 11.1\% & $-0.34$ vs $+0.11$ & 29\% \\
\bottomrule
\end{tabular}
\end{table}

Neither kind of limit saves much of the load. The adopted caps bind too few authors to matter, and the past-acceptance rules block few submissions: ICLR's rule blocks 3.6\% and saves about 4\% of the hours, and even the strict rule blocks 10\% and saves 10\%.

Nor do the limits improve who is accepted. The credential barely separates strong authors from weak, because the same noisy funnel issued it. After three cycles, the 47\% of active authors who hold a past acceptance exceed the rest in mean ability by only about a tenth of a standard deviation. So the papers the past-acceptance rules block are about as good as those they admit. And because the conference accepts as many papers as before, each good paper kept out is replaced by whatever the same three noisy reviews accept next. Under ICLR's rule, 89\% of top-3\% papers are accepted, against 91\% with no limit. The strict rule does worse: it keeps one top-3\% paper in ten from being submitted at all, and of all the top-3\% papers written it accepts 84\%. At the top, no limit changes much. Under each, a top-0.5\% paper becomes a spotlight or oral 34--38\% of the time, against 37\% with none.

Only the self-limit cuts the load much, and it pays in the best papers. It saves 24\% of the hours, and it does select: the papers authors drop are worse on average than those they keep ($-0.34$ against $+0.11$). But it drops prolific authors' third and later papers before anyone reads them, among them about 6\% of the top-3\% papers. Its benefit also rests on authors rating their own papers better than reviewers do: the model gives their ratings a noise of 1, against a full review's 2.45 (Appendix~\ref{app:authors}).

The two kinds also fall on different authors. Authors who entered in the latest cycle are about 29\% of the model's pool, yet they lead 68\% of the papers the strict rule blocks and 76\% of those ICLR's rule blocks, though in the model they arrive with the same spread of ability as everyone else. A newcomer's paper can qualify only through a co-author who already holds the credential, so the past-acceptance rules fall on new authors by construction, at every correlation. The self-limit falls on prolific authors instead: the newest authors lead 29\% of the papers it blocks, their share of the pool.

\section{Engineering the funnel}
\label{sec:design}

A remedy, then, has to change how the funnel spends its judgment: a little on every paper, to rank it, and full reviews only for the papers ranked near the top. Its readers must come from the submissions, so that they keep pace with the papers. Following decision-making engineering, we chose each part of such a funnel by simulating its alternatives, and the result is the staged funnel (Table~\ref{tab:volumes}). An optional AI gate turns back papers not ready for review; each submission names a peer, who reads for it; peers rank the papers that pass by sorting their gists; and full reviews check and rank only the finalists. Simulation also fixed most of the design's settings, and it shows why we recommend this design over a cheaper one that matches today: the cheaper design falls below today as soon as the reading is noisier than assumed, while ours keeps its lead.

\begin{table}[ht]
\caption{The staged funnel per 1,000 submissions. A sort, one peer ordering ten gists, takes about an hour; a full review takes two. Of the 245 accepted, 32 are spotlights or orals and about 4 are orals.}
\label{tab:volumes}
\centering
\footnotesize
\setlength{\tabcolsep}{3.5pt}
\begin{tabular}{lllrrrr}
\toprule
Stage & Who reads & What & Papers in & Papers out & Judgments & Expert hours \\
\midrule
AI gate & AI & full paper & 1,000 & 800 & 1,000 AI reviews & 0 \\
Gist stage & peers & sets of ten gists & 800 & 300 finalists & 1,600 sorts & 1,600 \\
Full reviews & peers & full paper & 300 & 245 accepted & 900 full reviews & 1,800 \\
\midrule
\multicolumn{5}{l}{Per submission, owed by its peer} & 1.6 sorts, 0.9 full reviews & 3.4 \\
\bottomrule
\end{tabular}
\end{table}

The AI gate reads each full paper and checks its form and basic soundness: whether it is clearly written and free of basic errors of reasoning. It does not judge the idea. The stage after it cannot do this work. Peers rank papers by their gists, so a badly written or unsound paper can rank well on its gist and reach the finalists, where its full reviews would be spent screening it out. This screening is the main reason for the gate, and the model cannot show it: in the model a gist shows a paper's quality just as the full paper does, so there the gate only saves hours. Without it, the papers it would have cut are sorted too, so the same reading per paper costs 3.8 hours per submission instead of 3.4, and places papers about as well (Table~\ref{tab:compare} in Appendix~\ref{app:results}). We therefore count the gate as the reasonable choice, but an optional one: a conference that will not let an AI reject papers can drop it, and the full reviews then take on the screening. Its pass rate of 80\% is a setting of the design, not a measured rejection rate; a running funnel tunes it (below).

The readers come from the submissions. Each submission names one of its authors as its \emph{peer}, who owes the reading that submission costs: on average about 1.6 sorts and 0.9 full reviews a cycle (Table~\ref{tab:volumes}). The corresponding author serves, or a co-author they designate, and the role needs no invitation, seniority or past acceptance. An author on several submissions owes a turn for each, but most papers have several authors, so co-authors share the turns and each stays within the readers' cap of two sorts a cycle (below). A sort is also light work: about an hour, against two for a full review. Because the duty is tied to the paper, there are enough peers, and they grow with the submissions.

The peers rank the papers that pass the gate by their \emph{gists}. The AI drafts each gist from its paper in a fixed template of four parts: the problem, the idea, the claimed result and the evidence offered for it. The authors correct and approve it in about 30 minutes. Because every gist puts the same things in the same places, peers compare like with like, while the evidence itself stays in the full paper for the finalists' full reviews. In a \emph{gist sort}, a peer orders ten gists, drawn at random, from best to worst, and writes no review; at our working figure of six minutes a gist, a sort takes about an hour. Each paper is sorted by twenty different peers, so no single peer decides where it stands, and its twenty places are combined into its \emph{gist score} (Appendix~\ref{app:model}). Peers elsewhere already rank such texts at scale. At the ALMA observatory, proposers rank one another's short proposals in sets of ten; over five cycles and more than 8,000 proposals, their rankings reproduced the panels' systematic trends, with a dispersion comparable to that of panel reviewers' independent scores before discussion \citep{alma2026}. ALMA made its review dual-anonymous at the same time as it moved to ranking by proposers, though, so the effects of the two changes cannot be separated.

We chose sorting over scoring, in which each peer reads one gist and gives it a score, by simulating both at equal hours. A score says a little more than a place among ten, but it takes about 15 minutes against six, so the hours that buy a paper twenty sorts buy it only eight scores. At equal hours, sorting places papers better: its hit rate at the line is 0.81, against 0.76 for scoring (Appendix~\ref{app:design-reading}). Sorting also cancels a peer's personal scale, which scoring carries into every score (Table~\ref{tab:leniency} in Appendix~\ref{app:results}).

The top 30\% of submissions by gist score become \emph{finalists}, and three peers review each in full, both to check it and to rank it. The check, which we call the \emph{delivery check}, asks whether a finalist delivers what its gist claims: whether the evidence supports the claims, the proofs and experiments are correct, and the novelty survives the prior work. A confirmed flaw removes the paper, and there are three reviews so that a majority can settle a disputed one. For the ranking, each finalist's gist score is combined with its full reviews, each weighted by the number of judgments behind it, twenty sorts against three reviews, and the combined ranking sets acceptance, spotlights and orals at today's rates. The gist score carries most of that ranking: ordered by their full reviews alone, the finalists would make a top-0.5\% paper a spotlight or oral 67\% of the time, against 97\% when the gist score is included (Table~\ref{tab:last} in Appendix~\ref{app:results}). Area chairs no longer rank papers but keep their other roles, such as adjudicating a reported flaw; neither these roles nor the delivery check is in the model.

In return for its peer's reading, every submission receives feedback wherever it stops. A paper cut at the gate receives the gate's report, which its authors can act on before resubmitting. The gate, if kept, turns back a fifth of submissions this way. A paper cut at the gist stage receives that report and its \emph{rank band}, the place its gist took on average in its sets of ten, such as fifth or sixth. Neither receives a written review, where today every paper gets three. That is the funnel's price to authors.

Simulating each part fixed most of the design (Appendix~\ref{app:design}). The structure sets the cost, since full reviews go only to finalists. The number of sorts per paper sets the placement, and what a peer can read limits it: we assume a peer can sort at most two sets of ten in a cycle, and twenty sorts per paper ask 1.6 sorts of each peer. Two settings depend on noise that only a running funnel can measure, so they are left to its data. The gate's pass rate trades sorts against good papers: with each peer's load held, a tighter gate gives every paper it passes more sorts, but it also cuts more papers that belong above the accept line. In the model, the hit rate at the line peaks when the gate cuts about one in twenty of those papers, and a gate that reads as well as one full review reaches that point at a pass rate of 80\%. A running funnel finds its own rate with an \emph{audit sample}: a random 5\% of the cut papers, about 300 a cycle at NeurIPS 2026's scale, go on to the gist stage, and where they rank shows roughly how many cut papers belonged above the line. Likewise, the agreement among each paper's sorts measures the gist sorts' noise, which sets the gist score's weight. The remaining choices express goals the model cannot weigh, so they belong to the organizers: whether to keep the gate, how many full reviews the delivery check needs, and how selective to be (Table~\ref{tab:design-record} in Appendix~\ref{app:design-record}).

A search over all settings together checks the design and shows why the cheapest design that matches today is not the one to adopt. That design costs 2.0 hours per submission, a third of today's, equals today's 0.63 at the line and reaches 0.36 at the top, where today's is 0.20. But it has no margin. If every peer judgment is twice as noisy as assumed, its hit rate at the line falls to 0.56, and if the gist sorts are three times as noisy, to 0.52, both below today's. Under the same stresses ours falls only to 0.75 and 0.71 (Table~\ref{tab:design-rates} in Appendix~\ref{app:design-search}). It lies where the search's frontier stops climbing steeply, past about three hours (Figure~\ref{fig:design-search}, in the same appendix), and its extra 1.4 hours buy the margin.

\section{What the staged funnel buys}
\label{sec:buys}

At 3.4 expert hours per submission, against 6 today, the staged funnel places papers better than today's funnel or any limit, and more reproducibly than today's funnel. It gains because it ranks each paper that passes the gate by twenty cheap judgments instead of three expensive ones. It has one weak spot: just below the spotlight and oral cuts, it promotes a few more papers than today's funnel does. We know the cause and which settings should ease it, though we have not yet simulated how far they do. And because a tighter acceptance rate lowers its placement only a little while freeing hours, a conference can set the rate by its goals.

At the accept line, the staged funnel accepts more of the papers that belong above it and fewer of those below (Figure~\ref{fig:placement}). A 3--10\% paper is accepted 95.5\% of the time instead of 74\%. A bottom-half paper is accepted 0.3\% of the time, so the bottom half supplies fewer than one accepted paper in a hundred, where today it supplies one in ten. The hit rate at the line rises from 0.63 to 0.81. The accepted list is also more reproducible: a rerun keeps about three-quarters of it (0.726), not half.

At the top, a top-0.5\% paper becomes a spotlight or oral 97\% of the time, against 37\% today, and a rerun keeps about half of the spotlights and orals (0.54) instead of a tenth. About half of that gain needs no new funnel: if today's chairs chose spotlights and orals by the reviewers' mean scores, a top-0.5\% paper would become one 66\% of the time (Table~\ref{tab:compare} in Appendix~\ref{app:results}). The rest of the gain at the top, and all of it at the accept line, comes from the staging.

Today's funnel could buy this placement only at great cost (Figure~\ref{fig:cost}). To accept a 3--10\% paper as often, it would need about fifteen full reviews per paper, or 30 hours. At the 2026 scale of 30,709 submissions, the staged funnel takes about 104,000 expert hours instead of 184,000, and so returns about 80,000 hours a cycle, or 40 researcher-years, to research (Table~\ref{tab:hours} in Appendix~\ref{app:hours}). Its full reviews also fall mostly on papers that end up accepted: 82\% of their hours, and 72\% on papers in the top quarter of quality. Today, when every paper gets the same three, those shares are only 24.5\% and 25\%.

\begin{figure}[t]
\centering
\includegraphics[width=\textwidth]{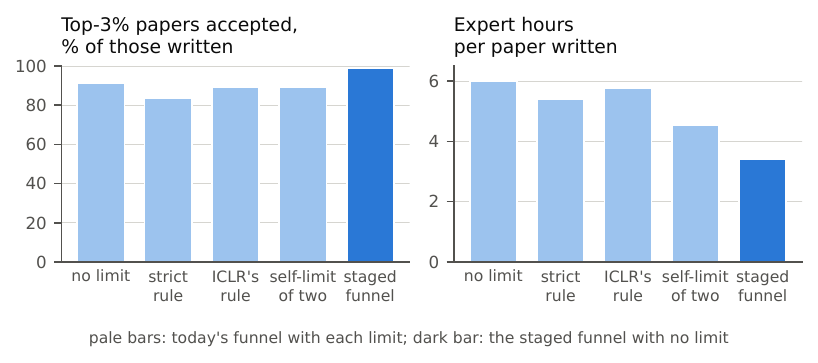}
\caption{The limits on who may submit, under today's funnel, against the staged funnel with no limit (computed in the model with authors, as in Table~\ref{tab:limits}, which gives the limits' numbers).}
\label{fig:limits}
\end{figure}

Against the limits, the staged funnel does better on both counts and blocks no one (Figure~\ref{fig:limits}). On Section~\ref{sec:limits}'s measure, the top 3\% counted against all papers written, it accepts 99\% of them at all three correlations, for 3.4 hours per paper written (Table~\ref{tab:authors} in Appendix~\ref{app:authors}); the limits accept 84--89\%. It also answers the reason ICLR gives for its new-author rule, a shortage of qualified reviewers: each submission brings a peer.

Just below the top cuts, though, the staged funnel promotes a few more papers than today's funnel does (Figure~\ref{fig:placement}; Table~\ref{tab:buys}, page~\pageref{tab:buys}). A 3--10\% paper, which belongs below the spotlight cut, becomes a spotlight or oral 15\% of the time, against 12\% today. A 0.5--3\% paper, which belongs below the oral cut, becomes an oral 5\% of the time, against 3\%. Each such promotion takes one of a fixed number of places from a paper above the cut. The cause is that the staged funnel's mistakes are near misses. Today's three reviews and the chairs' noise can lift a paper from far below a cut: 6\% of 10--25\% papers become spotlights or orals. The staged funnel ranks each paper by twenty judgments, so it seldom misjudges a paper by that much, and only 0.8\% of 10--25\% papers are promoted. The papers it promotes by mistake therefore come from just below the cut, where they differ least from the papers above it and readers agree least (Section~\ref{sec:intro}). That tier gains promotions; the tiers below lose them.

The settings to tune are those that sharpen the ranking at the top: the gist score's weight in the final ranking, more sorts per paper up to what a peer can sort in a cycle, and a \emph{highlight stage}, in which the top 5\% of submissions get six more full reviews before spotlights and orals are chosen. In the model, each moves the hit rate at the top, now 0.66 for the staged funnel against today's 0.20, and barely moves the one at the line (Appendices~\ref{app:design-trades} and~\ref{app:design-lever}). Those runs report only the hit rate at the top, so how far each setting reduces these promotions has not yet been simulated.

Under today's funnel, a tighter acceptance rate still costs six hours a submission, because every paper still gets three full reviews, and it leaves more of who gets in to chance: the hit rate at the line falls from 0.63 at 25\% to 0.51 at 10\%. Under the staged funnel it falls only from 0.81 to 0.75. With fewer finalists to review, going to 10\% also frees about 1.06 hours per submission, which at the 2026 scale comes to some 32,500 hours, or 16 researcher-years, a cycle (Table~\ref{tab:design-lever} in Appendix~\ref{app:design-lever}). A conference can keep those hours, spend them on more judgment at the top, give a written review to each paper between the 10\% and 25\% lines, or draw a main track and a second tier from one ranking. Which rate is right depends on what the conference is for.

\section{How much can go wrong}
\label{sec:risks}

The staged funnel's advantage survives large errors in each of its unmeasured assumptions, and a less favorable reading of the measured noise. What the model cannot see lies elsewhere: biases that every reader shares, peers and authors who game the funnel, and the low-effort submissions a flood may bring. Table~\ref{tab:risks} sets each stress beside the funnel as assumed and beside today's funnel.

\begin{table}[t]
\caption{The staged funnel under stress, beside today's funnel (computed; more stresses in Table~\ref{tab:stress}).}
\label{tab:risks}
\centering
\footnotesize
\setlength{\tabcolsep}{2pt}
\begin{tabular}{@{}lrrrrrrrr@{}}
\toprule
 & As assumed & Gist sorts $2\times$ & Gist sorts $3\times$ & Peers $2\times$ & Peers $6\times$ & Gate $8\times$ & Shared noise & Today \\
\midrule
Top 0.5\%: accepted & 99.7 & 99.9 & 99.8 & 99.8 & 98.2 & 93.4 & 99.7 & 98 \\
Top 0.5\%: spotlight or oral & 97 & 89 & 83.5 & 88 & 63 & 90 & 76 & 37 \\
3--10\%: accepted & 95.5 & 90.5 & 85.8 & 90.0 & 74.2 & 87.6 & 81.5 & 74 \\
Accepted list reproduces & 0.726 & 0.654 & 0.604 & 0.649 & 0.505 & 0.668 & 0.565 & 0.505 \\
\bottomrule
\end{tabular}
\end{table}

The largest bet is the gist sort. A gist says less than a paper, and a peer sorting ten of them in an hour may judge each paper less well than a full review does; how much less has not been measured. The margin is wide. With gist sorts twice as noisy as a full review, the staged funnel still accepts a 3--10\% paper 90.5\% of the time, and with gist sorts three times as noisy, 85.8\%.

Today's reviewers are selected; peers drawn from the submissions are not. But the bar they must clear is low, since in the model a single full review correlates only about 0.54 with quality. If every peer judgment, gist sorts and full reviews alike, has twice today's noise, the staged funnel still accepts 90\% of 3--10\% papers, and it falls to today's level only when peers are about six times as noisy.

The gate is the one part where a good paper can fare worse than today, and only if the gate is much noisier than a full review. A gate as reliable as one full review cuts 0.3\% of top-0.5\% papers (Table~\ref{tab:tiers} in Appendix~\ref{app:results}). A gate eight times as noisy lowers the share of top-0.5\% papers accepted to 93.4\%, below today's 98\%, while the funnel still accepts 87.6\% of 3--10\% papers. The pass rate of 80\% keeps these losses small, and the audit sample measures them every cycle.

The measured noise also allows a less favorable reading. The correlation of 0.55 cannot tell whether review noise is each reader's own, which more readers average away, or shared by everyone who reads a paper in a cycle, which no number of readers removes. The main runs assume the first. If instead part of the noise is shared, with the split set so that three full reviews still correlate at 0.55, a paper's twenty sorts average away less of it, and the gain shrinks but holds (Table~\ref{tab:risks}). Today's funnel, calibrated to the same measurement, places papers as before under either reading.

A measure built on readers' shared judgment cannot see a bias that every reader shares, such as presentation: polished figures, fluent prose, the look of an established lab. To see which way it would push, we gave papers a presentation effect apart from their quality, with half of quality's variance, and scored placement against quality alone (Table~\ref{tab:polish} in Appendix~\ref{app:results}). Today's funnel grows more reproducible and less right: its accepted list reproduces at 0.56, while it accepts a 3--10\% paper 66\% of the time, down from 75\% in the same runs without presentation. These runs assume a gist's fixed template strips presentation, so the staged funnel barely moves (94.6\%). They show which way the effect goes, not how large it is; how much presentation sways readers has not been measured.

Gaming lies outside the model, and the design's defenses against it are untested. A peer whose own paper competes with the gists in a set can gain by ranking them low; partition mechanisms, in which each peer ranks only papers outside their own group, remove that gain at some cost to the selection \citep{alon2011,aziz2016,xu2019}, and other impartial mechanisms remove it too \citep{kurokawa2015}. Authors who collude to rank one another's gists high face three obstacles: sets are drawn at random, each paper has twenty different sorters, and the sort data show how far each peer agrees with the consensus. An author who oversells a paper in its gist meets the delivery check.

The model also has no low-effort submissions. If the flood brings papers written with little effort, the gate is the part of the funnel built to stop them, and we expect a real gate to cut more than the fifth the design sets. That expectation is speculative, since no gate has been tested against such papers.

\section{Conclusion}
\label{sec:conclusion}

Conferences should change how their funnel judges papers, not who may submit them. We reached this recommendation through decision-making engineering, simulating each design before adopting it, in a model calibrated to the NeurIPS experiments. The simulations showed what the designs' rules alone do not show: the limits being adopted save little and place papers no better, and the cheapest design that matches today falls below it when peer judgments or gist sorts are noisier than assumed. They also found better funnels than today's. The staged funnel worked through here is one: for a little over half of today's expert hours, it places papers closer to where readers' shared judgment puts them, at the accept line and at the top. Its lead holds even when its largest assumption is far off: with gist sorts three times as noisy as a full review, it still accepts 85.8\% of 3--10\% papers, against 74\% today. Any rule a conference considers can be simulated the same way before it is adopted.

The change asks something of authors. Each submission owes, through its peer, the reading it costs, and a paper cut before the finalists, by the gate or at the gist stage, receives feedback but no written review, unless the conference spends hours freed by a tighter rate on one. In return, about 40 researcher-years a cycle go back to research at the 2026 scale, and the peers keep pace with the flood, since each new submission brings its own. New authors submit on the same terms as established ones, and a conference sets how selective to be by its goals, not by its reviewers' hours.

What remains is for a running funnel to measure. A conference that adopts the staged funnel can let its own data tune what we could only assume: the audit sample shows how often the gate cuts a paper from above the accept line, and the agreement among each paper's sorts, and between each peer and the consensus, shows how well peers judge gists. A running funnel must also show whether twenty sorts of short gists favour work that is easy to summarize and hard to dispute, the drift toward conservative work that distributed review has long been suspected of (Appendix~\ref{app:related-incentives}); the model cannot. No such data can settle whether a selection that follows readers' shared judgment is also right; that needs a measure of quality beyond agreement.

\clearpage
\bibliographystyle{plainnat}
\bibliography{references}

\appendix

\section{The model and its calibration}
\label{app:model}

\paragraph{Judgments.} Each paper has a quality $s \sim \mathcal{N}(0,1)$. Every judgment of a paper sees $x = s + \varepsilon$ with $\varepsilon \sim \mathcal{N}(0, v)$, independent across judgments; this holds alike for a full review, a place in a gist sort and the AI review. The mean of $k$ full reviews in two independent runs correlates at $1/(1 + v/k)$. Setting the three-review value to the measured 0.55 \citep{cortes2021} gives $v = 3(1/0.55 - 1) = 2.45$; a single full review then correlates $1/\sqrt{1+v} \approx 0.54$ with $s$. This differs from Cortes and Lawrence's own fit, whose Table~2 puts a single review's noise at about the variance of quality, 1.27 against 1.28, and which they read as consistent with the same 0.55 \citep{cortes2021}. We fit $v$ to 0.55 directly, as what it measures: the correlation of two committees' mean scores. The 2014 papers had about 3.6 reviews each rather than three, and with 3.6 the same fit would give $v \approx 2.9$.

\paragraph{Today's funnel.} Three full reviews are averaged and the top 24.5\% of submissions accepted. The chairs choose the 3.2\% spotlights or orals and the 0.36\% orals from the accepted papers by the mean score plus extra noise of variance $d$. These rates are NeurIPS 2025's: the accept rate from its program chairs \citep{neurips2025stats}, and the spotlight and oral rates from its 688 spotlights and 77 orals \citep{riken2025}, with the spotlight rate applied to spotlights and orals together; the 25.7\% that NeurIPS 2026 accepted is close \citep{riken2026}. The accept step has no chair, because in 2021 a cut on the reviewers' mean final scores reproduced the accepted list as well as the committees' own decisions did \citep[read from Fig.~3]{beygelzimer2023}. Counted over all decisions rather than over the accepted list, the 2021 committees differed on 23\% of papers, where random decisions would differ on 35\% \citep{beygelzimer2023}. Fitted over $d \in \{0, 0.25, 0.5, 1, 1.5, 2, 3\}$, $d = 2.0$ gives a reproducibility of 0.113 for the spotlights and orals, against the measured 0.11 \citep{beygelzimer2023}. Two results were not fitted. The accepted list reproduces at 0.505, against the measured 0.50; and with $d = 0$, spotlights and orals chosen by score, they reproduce at 0.236, against about 0.25 read from the same figure of \citet{beygelzimer2023}.

\paragraph{The staged funnel.} The AI gate sees $s$ with noise $v_{\mathrm{AI}}$ ($= v$ in the main runs) and passes the top fraction $g = 0.8$ of submissions. In each of $k_g = 20$ rounds, the papers that pass are shuffled into sets of ten, and one peer orders each set by judgments of noise $v$, drawn afresh for each paper in each set; a paper's place in the set gives it a normalized rank from 1 (best) to 0 (worst). Its gist score is the mean of its twenty ranks. The top fraction $f = 0.30$ of submissions by gist score become finalists, and each receives $m = 3$ full reviews of noise $v$. Among the finalists, with $z_g$ and $z_f$ the standardized gist score and full-review mean, the combined score is $(k_g z_g + m z_f)/(k_g + m)$; its rank sets acceptance, spotlights or orals, and orals at the same rates as today. There is no chair step. In option A, each passed paper instead receives $k_g = 10$ gist scores, and the two stages' means are pooled with weights $k_g/v_g$ and $m/v_f$.

\paragraph{Combining the sorts.} Three ways of combining a paper's twenty ranks give the same placement within simulation error (Table~\ref{tab:recon}): the mean rank, which the figures use; the mean normal score, in which each position is replaced by its expected standard-normal order statistic (Blom's approximation); and Bradley--Terry, which reads each sort as its 45 pairwise wins and fits strengths by a minorization--maximization iteration with a weak prior of one virtual win and one virtual loss. Because sets are drawn at random, they are balanced, and the three agree. Bradley--Terry is marginally best, and it corrects for a paper that lands in unusually strong or weak sets, which matters more when sets are not random, for example when they are drawn within a subfield. We recommend it; the mean rank is a simple fallback.

\paragraph{Leniency.} In the leniency runs, a scoring reader adds a personal offset of variance $L$, in units of quality's variance, to every score they give. A sorting reader's offset shifts every judgment in their set alike and so cancels. Cortes and Lawrence's calibration of the 2014 scores puts a reviewer's offset at variance 0.24 against 1.28 for paper quality \citep[Table~2]{cortes2021}, about a fifth of quality's variance; the runs show $L$ at 0, 0.5, 1 and 2.

\paragraph{Presentation.} In the presentation runs, each paper also has a presentation effect $p \sim \mathcal{N}(0, P)$, independent of $s$. Every full review and the AI gate see $s + p$ plus noise; a judgment in a gist sort sees $s$ plus noise. Placement is scored against $s$. The review noise is held at 2.45, so the runs show the direction of the effect, not its size: with $P > 0$, today's funnel would need recalibrating to the measured 0.55, because presentation adds variance that is stable across reruns. $P$ is not measured; it is shown at 0, 0.25 and 0.5 (Table~\ref{tab:polish}). At $P = 0.5$, today's accepted list reproduces at 0.56 instead of 0.50 while it accepts a 3--10\% paper 66\% of the time instead of 75\%; the staged funnel's figures barely move.

\paragraph{Shared noise.} In the main runs every judgment carries its own noise. Under \emph{shared} noise, every judgment carries noise 1, plus a per-paper error of variance 0.485 that every human judgment of the paper in one run shares, chosen so that three full reviews still correlate at 0.55. Under shared noise the staged funnel accepts 81.5\% of 3--10\% papers, against 74\% today, and its accepted list reproduces at 0.565 (Table~\ref{tab:stress}). Gist noise is varied by multiplying the noise of judgments within sorts; peer noise by multiplying that of sorts and full reviews together; gate noise by multiplying $v_{\mathrm{AI}}$.

\paragraph{Hours.} Today's funnel costs $2k$ expert hours per submission. The staged funnel costs $g\,k_g\,t + f\,m \cdot 2$, with $t = 0.1$ hours (six minutes) per gist in a sort, and option A $g\,k_g \cdot 0.25 + f\,m \cdot 2$. An audit sample of a fraction $a = 0.05$ of the cut papers, each sorted $k_g$ times, adds $(1-g)\,a\,k_g\,t = 0.02$ hours. The chairs' time and the AI are excluded on both sides, and the authors' time approving their gists is not counted as review.

\paragraph{Simulation.} Each run draws 20,000 papers. Each result pairs two independent runs on the same papers, which gives reproducibility directly. The main tables use 40 paired runs and the sweeps 12 to 20. Placement percentages carry a simulation error of about $\pm$0.5 points in the large tiers and $\pm$1.5 points in the top-0.5\% tier, and reproducibility about $\pm$0.005, so the same quantity can differ by a point between tables drawn from different runs.

\paragraph{Peers at ALMA.} The ALMA comparison is measured, not computed \citep{alma2026}. Over Cycles 8 to 12, 8,086 proposals and 80,814 reviews, proposers ranking one another's proposals in sets of ten reproduced the panels' systematic ranking trends, with a dispersion comparable to that of panel reviewers' independent scores before discussion. The switch to distributed review coincided with the introduction of dual-anonymous review, so the two effects cannot be separated.
\section{Designing the funnel}
\label{app:design}

\subsection{The question}
\label{app:design-question}

The paper's design (Section~\ref{sec:design}) leaves an organizer fair questions. Why twenty sorts per paper and not five? Why an AI gate that passes 80\% and not 50\%? Why finalists at 30\% of submissions when a quarter are accepted, and why three full reviews for each? How should the gist score and the full reviews be combined? Why sort, if scoring places papers almost as well? And why accept a quarter at all? Without the answers, the design looks picked. An organizer cannot tell which of its settings the numbers force and which are free, and so cannot adapt it to a venue with other goals. We retrace the design one choice at a time, computing what each choice costs in hours and buys in placement before we make it. This is the paper's decision-making engineering (Section~\ref{sec:intro}), turned on its own design.

The design is a set of priced choices of four kinds. The first is structure: giving full reviews only to finalists sets the cost. The second is the reading: how many gists each reader sorts sets the placement, up to a cap on what a reader can do, and within that cap, sorting sets of ten is the best use of a reader's hour. The third is two trades that a running funnel's own data settle: how many papers the gate passes, and how much weight the final ranking gives the gist score. A search over these choices together checks the paper's design, and shows why the cheapest design that matches today is not the one to adopt. The fourth kind belongs to the organizers: whether to use the gate; three full reviews per finalist and a margin of extra finalists, which buy something the model cannot see; and the acceptance rate. The gate is part of the design and the reasonable choice, but it is optional: without it, the full reviewers screen out the badly written or unsound papers it would have cut, at the cost of their hours. Under the staged funnel the acceptance rate is a lever: a broad accepted list holds fewer weak papers than today's selective one, a selective list frees hours, and one ranking can yield both.

\subsection{Hours and placement}
\label{app:design-hours}

Each choice below is priced in expert hours per submission and in placement, and giving full reviews only to finalists is where the saving in hours comes from. In the model of Appendix~\ref{app:model}, a gist in a sort takes six minutes and a full review two hours. Every submission names one reader, who owes the reading the submission costs (Section~\ref{sec:design}), so the reading grows with the submissions. A submission's expert hours are
\begin{equation*}
\text{hours} \;=\; \underbrace{\text{pass rate} \times \text{sorts per paper} \times 0.1}_{\text{gist reading}} \;+\; \underbrace{\text{finalists} \times \text{full reviews per finalist} \times 2}_{\text{full reviews}},
\end{equation*}
with finalists as a share of submissions. Since sorts per paper are gists per reader divided by the pass rate, the gist reading comes to six minutes for each gist a reader sorts, whatever the gate passes. Today's 6~h is three full reviews of every paper. The paper's design spends 1.6~h on gist reading and 1.5~h on full reviews of the accepted quarter. The finalists beyond the accepted list are the \emph{finalist margin}: 20\% of the accepted list in the paper's design, and their full reviews cost 0.3~h. The gate costs no expert time, so the total is 3.4~h. The 1.5~h is a floor, given three full reviews per finalist: the acceptance rate times three reviews times two hours. The structure sets the form of that floor, which falls with the acceptance rate (Appendix~\ref{app:design-lever}); the three in it is a judgment (Appendix~\ref{app:design-judgments}). The reading and the finalist margin are what the next choices spend.

Placement is judged at a fixed acceptance rate, starting at 25\%, with spotlight or better at 2.5\% of submissions and oral at 0.5\%. These are simplified from NeurIPS 2025's 3.2\% and 0.36\% \citep{riken2025} for easy arithmetic; the paper's own results use NeurIPS's rates. Two measures carry the walk. The \emph{hit rate at the line} is the share of the papers truly above the accept line that are accepted. The \emph{hit rate at the top} is the share of the true top 2.5\% that become spotlights or orals. Unlike the paper's quality tiers, drawn for an accept line near 25\%, the hit rate at the line works at any acceptance rate. Because the number accepted is fixed, it also counts both errors: each paper above the accept line that a design misses lets in one from below. Today's funnel scores 0.63 at the line and 0.20 at the top; the paper's design scores 0.81 and 0.66. At the paper's rates, that rise takes a 3--10\% paper's acceptance from 74\% to 95.5\%, and a top-0.5\% paper's chance of a spotlight or oral from 37\% to 97\% (Section~\ref{sec:buys}). We also report reproducibility, the share of the accepted list that a rerun with fresh readers accepts again (Section~\ref{sec:model}), because the paper's case rests partly on it (0.51 today against 0.73); no choice here turns on it.

\subsection{The reading}
\label{app:design-reading}

\begin{figure}[t]
\centering
\includegraphics[width=0.92\textwidth]{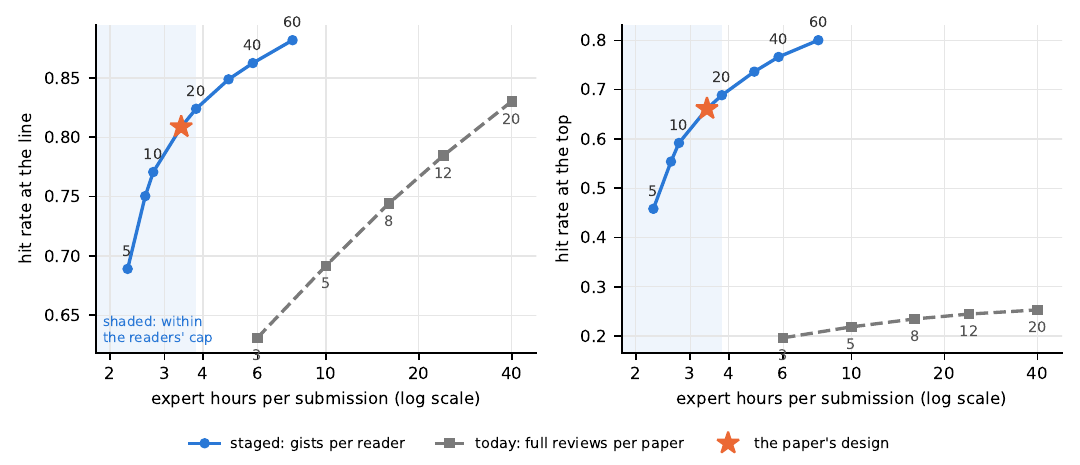}
\caption{What reading buys, at an acceptance rate of 25\% (computed). Hit rate at the line (left) and at the top (right) against expert hours per submission, on a log scale. Circles: the staged funnel as gists per reader rise from 5 to 60 (labels), with the gate passing 80\% and sets of ten; the shaded band lies within the readers' cap of 20 gists. Squares: today's funnel with 3 to 20 full reviews per paper (labels). Star: the paper's design, 16 gists per reader. The star's rise over today's three full reviews (0.63 to 0.81 at the line, 0.20 to 0.66 at the top) is, at the paper's rates, a 3--10\% paper's acceptance rising from 74\% to 95.5\% and a top-0.5\% paper's chance of a spotlight or oral rising from 37\% to 97\% (Section~\ref{sec:buys}).}
\label{fig:design-reading}
\end{figure}

The reading is where the design buys its placement, through three choices: how many gists each reader sorts, how large a set is, and whether readers sort or score. The \emph{readers' cap} bounds all three: a set holds at most ten gists, and a reader sorts at most two sets a cycle. Sets of ten follow ALMA \citep{alma2026}; two sets is our working assumption. Each choice is varied around the paper's design, with the others held.

How many gists should each reader sort? Nearly as many as the readers' cap allows. Placement climbs steeply with gists per reader and then flattens (Figure~\ref{fig:design-reading}): one set of ten per reader gives 0.77 at the line, the cap of two sets 0.82, and sixty gists, three times the cap, would give 0.88. The cap therefore decides how much placement the funnel can buy; a third set would give 0.85, at 4.8~h. Within the cap, the paper's 16 gists are a trade: against the cap's 20, they save 0.4~h and give up about 1.5 points at the line and 3 at the top. An organizer may take the 20. Today's funnel can buy placement only with more full reviews per paper: about fifteen, or 30~h, match the paper's design at the line, and twenty reach 0.25 at the top, at 40~h, because the chairs choose spotlights and orals with a noise of their own that more reviews do not remove (Appendix~\ref{app:model}).

How large should a set be? As large as we assume a reader can reliably sort: ten, the size ALMA's proposers sort. At a fixed load, a larger set gives each read more comparisons. With ten gists per reader, sets of ten beat sets of five by two points at the line and five at the top. With twenty gists per reader, they beat them by one point at the line and four at the top (Table~\ref{tab:design-choices}). Sets of twenty would add a point at the line, but we do not assume a reader can order more than the ten that ALMA's proposers sort.

\begin{table}[t]
\caption{One choice at a time, at an acceptance rate of 25\% (computed). Here and throughout this appendix, unless a caption says otherwise, numbers come from 12 paired runs of 20,000 papers, with a simulation error of about half a point on a hit rate. Each block varies one choice around the paper's design (gate passing 80\%, 16 gists per reader in sets of ten, 20 sorts per paper, finalist margin 20\%, three full reviews per finalist, merge weight 0.87), with the others held. Hours are expert hours per submission. Leniency, some readers scoring every paper high and others every paper low, is set at about a fifth of quality's variance, as estimated by \citet[Table~2]{cortes2021}; a sort cancels it. The row with each read in a sort 0.75 times as noisy assumes that ordering is easier than scoring, which no one has yet observed. The finalist margin moves the hit rate at the line by less than a point, so it is given to three decimals there. At acceptance rates of 10\% and 5\%, the best merge weight is also about 0.75, and weighting by judgments comes within half a point of it; with gists twice as noisy, weighting by judgments loses 1--2 points at the top there.}
\label{tab:design-choices}
\centering
\footnotesize
\setlength{\tabcolsep}{4pt}
\begin{tabular}{llrrr}
\toprule
Choice & Setting & Hours & Line & Top \\
\midrule
Sort or score & Sort, 20 per paper (the paper's) & 3.4 & 0.81 & 0.66 \\
 & Score, 8 per paper (equal hours) & 3.4 & 0.76 & 0.57 \\
 & \quad with leniency & 3.4 & 0.75 & 0.55 \\
 & Score, 20 per paper (equal reads) & 5.8 & 0.83 & 0.70 \\
 & \quad with leniency & 5.8 & 0.82 & 0.69 \\
 & Sort, each read $0.75\times$ as noisy & 3.4 & 0.83 & 0.69 \\
\midrule
Set size, at a & 10 gists, sets of 5 & 2.8 & 0.75 & 0.54 \\
fixed number of & 10 gists, sets of 10 & 2.8 & 0.77 & 0.59 \\
gists per reader & 20 gists, sets of 5 & 3.8 & 0.81 & 0.65 \\
 & 20 gists, sets of 10 & 3.8 & 0.82 & 0.69 \\
 & 20 gists, sets of 20 (beyond the cap) & 3.8 & 0.83 & 0.71 \\
\midrule
Finalist margin & 0 & 3.1 & 0.800 & 0.66 \\
 & 10\% & 3.25 & 0.808 & 0.66 \\
 & 20\% (the paper's) & 3.4 & 0.809 & 0.66 \\
 & 50\% & 3.85 & 0.809 & 0.66 \\
 & 100\% & 4.6 & 0.810 & 0.66 \\
\midrule
Merge weight on & 0 (full reviews alone) & 3.4 & 0.78 & 0.39 \\
the gist score, & 0.25 & 3.4 & 0.80 & 0.53 \\
as assumed & 0.5 & 3.4 & 0.81 & 0.64 \\
 & 0.75 & 3.4 & 0.81 & 0.67 \\
 & 0.87, by judgments (the paper's) & 3.4 & 0.81 & 0.66 \\
 & 1 (gist score alone) & 3.4 & 0.80 & 0.64 \\
\midrule
Merge weight, & 0 & 3.4 & 0.75 & 0.40 \\
gists twice & 0.5 & 3.4 & 0.76 & 0.59 \\
as noisy & 0.75 & 3.4 & 0.76 & 0.58 \\
 & 0.87, by judgments & 3.4 & 0.75 & 0.56 \\
 & 1 & 3.4 & 0.74 & 0.53 \\
\bottomrule
\end{tabular}
\end{table}

Should readers sort or score? Sort, because at equal hours sorting places papers better: 0.81 at the line against 0.76 for scoring (Table~\ref{tab:design-choices}). A score says a little more than a place among ten, but it takes two and a half times as long, so in the same hours a paper gets 20 sorts or 8 scores. At equal reads scoring is slightly better, at 0.83, but costs 5.8~h. Two further effects favor sorting. The first is \emph{leniency}: some readers score every paper high and others every paper low. Leniency enters every score and cancels in a sort. At the size Cortes and Lawrence's calibration gives it, about a fifth of quality's variance \citep[Table~2]{cortes2021}, it costs scoring a point at the line (0.76 to 0.75; Table~\ref{tab:leniency} shows more leniency). The second effect is untested: if ordering ten gists is easier than scoring one, so that each read in a sort is 0.75 times as noisy, sorting gains two more points at the line, to 0.83. A first cycle can test this (Appendix~\ref{app:design-record}). Sorting's advantage is mostly time, then, and time is what the funnel holds fixed.

\subsection{The trades a running funnel settles}
\label{app:design-trades}

\begin{figure}[t]
\centering
\includegraphics[width=0.92\textwidth]{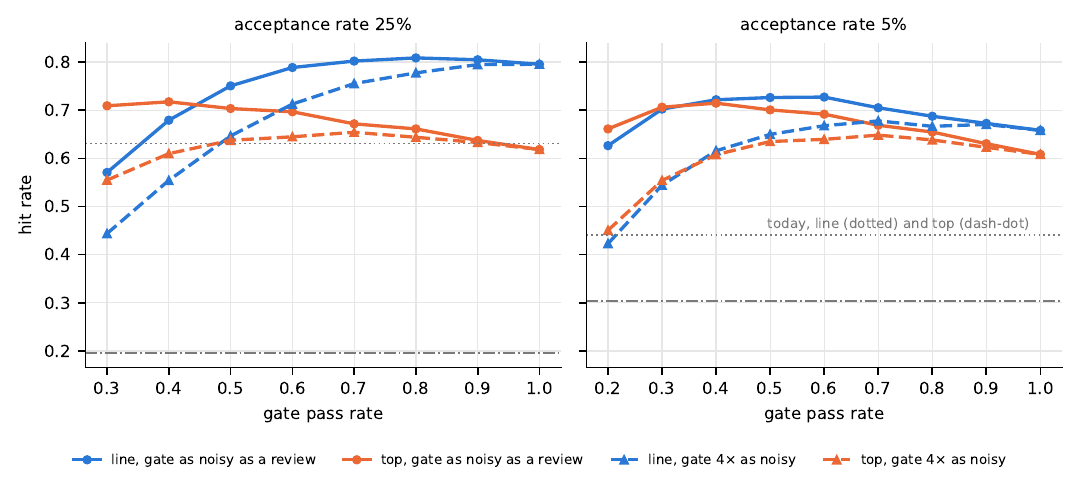}
\caption{What the gate's pass rate buys, with each reader's load held at 16 gists (computed). Hit rate at the line (blue) and at the top (orange) against the pass rate, at acceptance rates of 25\% (left) and 5\% (right). Solid lines: a gate as noisy as one full review; dashed lines: a gate four times as noisy. Grey lines: today's funnel at the line (dotted) and at the top (dash-dot). The audit's rough count, at NeurIPS 2026's scale and a pass rate of 80\%: if the gate cuts one in twenty of the papers above the accept line, about 19 of the 300 audited papers should rank above it; if one in ten, about 38. The audit sees a cut paper's rank among the sorts, not its quality, so some land on the wrong side, but 19 and 38 are still far enough apart to tell. Below the pass rate at which the hit rate at the line peaks, a tighter gate trades the hit rate at the line for the hit rate at the top. At 10\% (not shown), the hit rate at the line is best at a pass rate of 60--70\% for a gate as noisy as one full review and at 90\% for one four times as noisy.}
\label{fig:design-gate}
\end{figure}

The gate is optional, and using it is the reasonable choice (Section~\ref{sec:design}). Peers sort gists, so a badly written or unsound paper can rank well on its gist and reach the finalists. Without a gate, the full reviewers screen out such papers at the cost of their hours, a cost the model does not count, since in it a gist shows a paper's quality as the full paper does (Appendix~\ref{app:model}). Two settings have a best value that turns on a noise only a running funnel can measure: the gate's pass rate, where there is a gate, turns on the gate's noise, and the merge weight on the gists'. The \emph{merge weight} is the weight the final ranking puts on the gist score against the full reviews. For each setting, we give a rule that follows the measurement.

What a gate buys depends on what is held fixed. The paper holds sorts per paper at twenty; then the gate saves 0.4~h, about an eighth of the hours, for the same placement. Here each reader's load is held at 16 gists, as in Figure~\ref{fig:design-gate}; then the gate concentrates the reading on fewer papers, and it adds a point at the line (0.81 against 0.80) and four at the top (0.66 against 0.62). Against those numbers stands a question of policy: some conferences will not let an AI cut papers at all.

Where should the pass rate be set? Where the gate cuts about one paper in twenty from above the accept line. A tighter gate gives each surviving paper more sorts and keeps raising the hit rate at the top, but it cuts more papers from above the accept line. At 25\%, with a gate as noisy as one full review, the hit rate at the line peaks at a pass rate of 80--90\%. At 80\% the gate cuts 4\% of the papers above the accept line; at 50\% it cuts 22\%. But the best pass rate depends on the gate's noise, which no one knows before the gate runs. A gate four times as noisy is best at 90\% or fully open; at 80\% it would cut one paper in ten from above the accept line. A rule stated in the gate's errors carries across both, and in the model this one comes within about a point of the best hit rate at the line at every acceptance rate and gate noise we computed.

The audit sample estimates the share the gate cuts from above the accept line: the cut papers it sends to the gist stage are sorted with the rest, and where they rank shows how many belonged above the accept line. At NeurIPS 2026's scale it sorts about 300 of some 6,100 cut papers a cycle, roughly enough to tell one in twenty from one in ten (Figure~\ref{fig:design-gate} gives the count). We have not simulated the estimate. A conference that uses the gate can start it loose and tighten it until the audit shows about one in twenty cut from above the accept line. If the gate reads as well as one full review, as the paper assumes, the pass rate settles at 80\%; a gate four times as noisy would settle at 90\% or open. We expect a real gate to cut more than a fifth of submissions, since a flood may bring papers written with little effort, but that expectation is speculative (Section~\ref{sec:risks}).

How should the two stages be combined? By the evidence behind each. The paper weights the gist score by its share of the judgments behind the final ranking, 20 sorts against three full reviews, or 0.87; we call this \emph{weighting by judgments}. It matches the best weight, about 0.75, at the line, and comes within a point of it at the top (Table~\ref{tab:design-choices}). Full reviews alone lose the hit rate at the top: with no weight on the gist score, it falls to 0.39, as the paper found when it ordered finalists by their full reviews (Table~\ref{tab:last}). If gists are twice as noisy as assumed, the best weight falls to about 0.5, and weighting by judgments then costs a point at the line and three at the top. Weighting by judgments treats every judgment as equally precise; weighting by precision needs the gists' noise, which a running funnel measures every cycle in the agreement among each paper's sorts. So a funnel can start by weighting by judgments and follow precision after its first cycle.

\subsection{Checking the whole}
\label{app:design-search}

The paper's design was made one choice at a time, as above, and a search over the choices together checks it. The check holds: the paper's design lies where more hours stop buying much, and its gain over today survives readers noisier than assumed. The search also shows why the cheapest design that matches today is not the one to adopt, and that a search finds a region, not a setting.

\begin{figure}[t]
\centering
\includegraphics[width=0.92\textwidth]{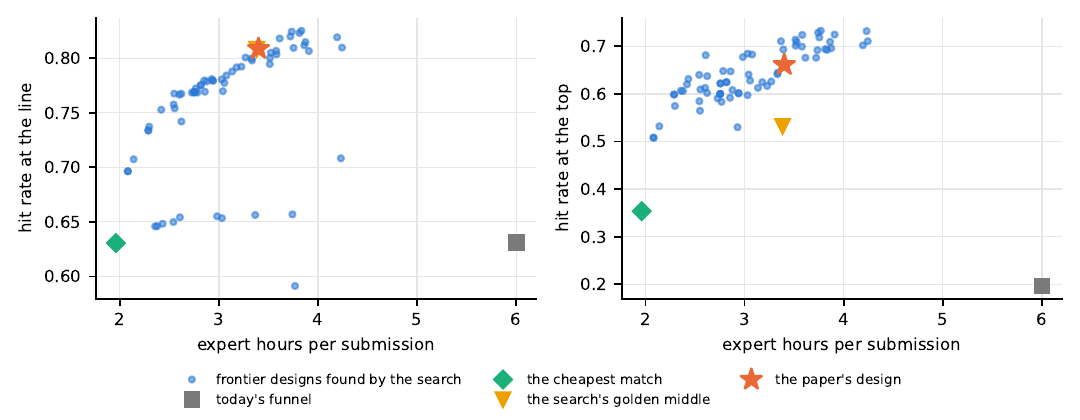}
\caption{The search at an acceptance rate of 25\% (computed). Hit rate at the line (left) and at the top (right) against expert hours per submission. Dots: designs on the search's frontier over hours and both hit rates, each scored on 3 shared runs; they scatter, and a dot that looks poor in one panel may be on the frontier through the other measure. Along the frontier, the hit rate at the line reaches 0.70 at 2.1~h, 0.80 at 3.3~h, and 0.82 at both 3.7~h and 4.2~h. Marked designs: today's funnel (square); the cheapest match (diamond) and the design the golden-middle rule picks (triangle, ``the search's golden middle'' in the legend), both scored again on 30 fresh runs; and the paper's design (star). The rule's design passes 76\% of papers and gives each reader two sets of nine gists (24 sorts per paper), with a reproducibility of 0.73. The search used Optuna \citep{akiba2019optuna} over the pass rate (0.3--1.0), set size (5--10), sets per reader (1--2), finalist margin (0--1) and merge weight (0--1), with three full reviews per finalist held. It found the cheapest match by a constrained TPE search \citep{bergstra2011tpe} and the frontier by NSGA-II \citep{deb2002nsga2}, 400 trials each.}
\label{fig:design-search}
\end{figure}

One choice at a time misses interactions, of the gate with the load and of the load with the finalist margin. So we searched five settings together within the readers' cap, with Optuna \citep{akiba2019optuna}: the pass rate, the set size, the sets per reader, the finalist margin and the merge weight. We held three full reviews per finalist, as a judgment (Appendix~\ref{app:design-judgments}); the caption of Figure~\ref{fig:design-search} gives the details. We scored the designs the search chose again on fresh runs, because the best result of a noisy search looks better than it is. The first thing the search finds is the \emph{cheapest match}: the cheapest design at least as good as today on both hit rates. It passes 92\% of papers, gives each reader one set of five gists, and has no finalist margin. It costs 2.0~h, a third of today's hours, equals today at the line (0.63), and does better at the top (0.36 against 0.20).

\begin{table}[t]
\caption{The paper's design and its reruns at other acceptance rates, against the cheapest match (computed). (a) At an acceptance rate of 25\%, under stress: hit rate at the line / at the top, with one noise variance multiplied at a time (peers: every peer judgment, gist sorts and full reviews alike; gists: judgments within sorts; gate: the AI review). The stresses were applied to these two designs, not to the design the golden-middle rule picks. Today's funnel has no gists and no gate, and under ``peers'' its reviewers stay as measured, because the stress asks whether the funnel's unselected peers read worse than today's selected reviewers (Section~\ref{sec:risks}). With its own reviewers twice as noisy, today's hit rate at the line would be 0.55 at 25\% (0.41 at 10\%, 0.33 at 5\%): if every reader got worse, the cheapest match, at 0.56, would tie today, and the paper's design, at 0.75, would still lead it by 20 points. (b) At three acceptance rates: the paper's design at 25\%, and the design at 10\% and the design at 5\%, each keeping 16 gists per reader in sets of ten, finalists 1.2 times the accepted and weighting by judgments, with the pass rate tightened. At 10\% the one-in-twenty rule alone gives 70\%; 60\% gives the same hit rate at the line (0.75) and more at the top (0.69 against 0.67). At 5\% the rule alone gives 60\%; the hit rate at the line is flat from 40\% to 60\% (0.72--0.73), and 40\% gives more at the top (0.72 against 0.69). Beside each design: today's funnel, the hours of the cheapest match, and the design's hit rate at the line with peers twice, gists three times and the gate four times as noisy. The designs the golden-middle rule picks at 10\% and 5\% (2.2--2.3~h and 1.9--2.0~h) match the design at 10\% and the design at 5\% at the line within a point.}
\label{tab:design-rates}
\centering
\footnotesize
\setlength{\tabcolsep}{3pt}
\begin{tabular}{lccccc}
\multicolumn{6}{l}{(a) Under stress, at 25\%} \\
\toprule
 & As assumed & Peers $2\times$ & Gists $2\times$ & Gists $3\times$ & Gate $4\times$ \\
\midrule
The paper's design (3.4~h) & 0.81 / 0.66 & 0.75 / 0.55 & 0.75 / 0.56 & 0.71 / 0.49 & 0.78 / 0.64 \\
The cheapest match (2.0~h) & 0.63 / 0.36 & 0.56 / 0.29 & 0.56 / 0.30 & 0.52 / 0.23 & 0.63 / 0.36 \\
Today (6~h) & 0.63 / 0.20 & & & & \\
\bottomrule
\end{tabular}

\vspace{1em}
\begin{tabular}{lccccc}
\multicolumn{6}{l}{(b) At three acceptance rates} \\
\toprule
Rate & Today & Design & Design & Cheapest & Design's line, \\
 & line / top / repro. & pass, sorts, hours & line / top / repro. & match & peers / gists / gate \\
\midrule
25\% & 0.63 / 0.20 / 0.51 & 80\%, 20, 3.4~h & 0.81 / 0.66 / 0.73 & 1.96~h & 0.75 / 0.71 / 0.78 \\
10\% & 0.51 / 0.25 / 0.36 & 60\%, 27, 2.3~h & 0.75 / 0.69 / 0.65 & 1.07~h & 0.68 / 0.64 / 0.69 \\
5\%  & 0.44 / 0.30 / 0.28 & 40\%, 40, 2.0~h & 0.72 / 0.72 / 0.61 & 0.77~h & 0.66 / 0.62 / 0.62 \\
\bottomrule
\end{tabular}
\end{table}

Why not adopt the cheapest match, at 1.4~h less? Because at the line it is level with today and has no margin there. Under stress, with one kind of reading noisier than assumed, it falls below today at the line (Table~\ref{tab:design-rates}): to 0.56 with peers twice as noisy and to 0.52 with gists three times as noisy, against today's 0.63. Its hit rate at the top stays above today's (0.23 at worst, against 0.20). The paper's design stays above today at the line under each stress: 0.75 with noisier peers, 0.71 with noisier gists, and 0.78 with a gate four times as noisy. The extra 1.4~h buys that margin.

The frontier is the set of designs that no other design beats on hours and both hit rates together. It climbs steeply to about 3~h and flattens after (Figure~\ref{fig:design-search}). We call the part where it flattens, and where a design's gain over today survives stress, the \emph{golden middle}. The paper's design, at 3.4~h and 0.81, lies there. To see whether a search left to itself would land there too, we gave it a golden-middle rule: take the cheapest frontier design with 90\% of the largest gain at the line and a hit rate at the top no worse than today's. The rule's design lands at the paper's 3.4~h and 0.81, and moving the 90\% threshold moves it along the flat part and no further: 80\% gives 3.1~h, and 95\% gives 3.6~h. It differs from the paper's design in two settings that barely move the hit rate at the line: a finalist margin of 4\% instead of 20\%, and a merge weight of 0.25 instead of 0.87. The merge weight costs the hit rate at the top 13 points (0.53 against 0.66), since the rule asked only that the hit rate at the top be no worse than today's. A search thus finds a region, not a setting. A setting that barely moves the hit rate at the line is left to the measurement that settles it, as the merge weight is (Appendix~\ref{app:design-trades}), or to judgment, as the margin is (Appendix~\ref{app:design-judgments}). And a rule for picking a design from the frontier has to name every measure the designer cares about.

\subsection{Two judgments the model cannot price}
\label{app:design-judgments}

Three full reviews per finalist and the finalist margin cost hours that the model counts, but buy something it cannot see, so it cannot weigh one against the other. They are now the largest cost the model does not justify. Each full review per finalist costs 0.6~h at 25\%, where finalists are 30\% of submissions; yet in the model one or two reviews place papers as well as three (Table~\ref{tab:last}). The finalist margin costs 0.15~h for every 10\% of the accepted list, and it buys about a point at the line up to 10\% and nothing beyond (Table~\ref{tab:design-choices}). What both buy lies in the delivery check, and no paper in the model fails it. Three reviewers let a majority settle a disputed flaw, and a margin leaves room to remove a finalist whose paper does not deliver what its gist claims. That is the judgment the paper makes: three reviews per finalist, which cost 1.2~h more than one would, and a 20\% margin, which costs 0.3~h. A first cycle measures what they buy: how often full reviewers remove a finalist, and how often a finalist's full reviews disagree with its gist rank. Those two counts would revise both settings.

\subsection{The acceptance rate, a lever}
\label{app:design-lever}

Under the staged funnel, the acceptance rate is a lever that a conference can set for its goals. A broad accepted list already holds fewer weak papers than today's selective one, a selective list frees hours, and one ranking can carry both accept lines.

The rate sets how many papers are accepted, and nothing else the conference showcases. The 25 spotlights and orals per thousand submissions, 5 of them orals, stay at every rate, so at 10\% a quarter of the accepted list is highlighted, and at 5\% half would be (Table~\ref{tab:design-lever}). Under today's funnel a tighter rate costs the same 6~h, since every paper still gets three full reviews, while chance decides more of who is accepted: the hit rate at the line drops from 0.63 at 25\% to 0.51 at 10\% and 0.44 at 5\%. Under the staged funnel the full-review floor falls with the rate, to 0.6~h at 10\% and 0.3~h at 5\%. Rerun at those rates with the gate tightened to 60\% and 40\% (Table~\ref{tab:design-rates} gives why), the design at 10\% costs 2.3~h and the design at 5\% 2.0~h, and they hold the hit rate at the line at 0.75 and 0.72. A tighter rate turns away the papers between the two accept lines: the staged funnel by quality, today's funnel largely by chance.

\begin{table}[t]
\caption{The acceptance rate as a lever, per 1,000 submissions (computed). The 25\% options accept 250 papers and the 10\% options 100; every option has 25 spotlights or orals, 5 of them orals. ``Orals'': the hit rate for orals, the share of the true top 0.5\% made orals. ``Top-10\% held'': how many of the 100 papers in the true top 10\% the accepted list holds. ``Top quarter'' and ``bottom half'': the share of the accepted list that comes from the true top 25\% and the true bottom 50\%. ``Staged, 10\%, keep the hours'' is the design at 10\%. The rows below it spend the freed hours: on reading up to the readers' cap (20 gists per reader); on six full reviews per finalist instead of three; on a highlight stage, in which the top 5\% of submissions get six more full reviews before spotlights and orals are chosen; on both the reading and the highlight stage; or on one written review for each paper between the 10\% and 25\% accept lines, which leaves placement unchanged. Two tiers: one ranking cut at the 10\% accept line for a main track of 100 papers and at the 25\% accept line for a second tier of 150; the hit rate at the line, reproducibility and make-up are the main track's, and ``held'' gives the main track and both tiers together. In ``Two tiers, today'', the hit rate at the line, reproducibility and make-up are today's 10\% list. Its top differs from ``Today, 10\%'' (0.20 against 0.25) because the chairs choose spotlights and orals from all 250 accepted papers in two tiers, but from 100 in a 10\% conference; its orals are 0.12 in both.}
\label{tab:design-lever}
\centering
\footnotesize
\setlength{\tabcolsep}{3pt}
\begin{tabular}{lrrrrrrrr}
\toprule
Option & Hours & Line & Top & Orals & Repro. & Top-10\% & Top & Bottom \\
 & & & & & & held & quarter & half \\
\midrule
Today, 25\% & 6.0 & 0.63 & 0.20 & 0.12 & 0.51 & 80 & 63\% & 10.5\% \\
Today, 10\% & 6.0 & 0.51 & 0.25 & 0.12 & 0.36 & 51 & 81\% & 3.6\% \\
\midrule
Staged, 25\% (the paper's) & 3.40 & 0.81 & 0.66 & 0.57 & 0.73 & 97 & 81\% & 0.7\% \\
\midrule
Staged, 10\%, keep the hours & 2.34 & 0.75 & 0.69 & 0.61 & 0.65 & 75 & 99\% & 0 \\
\quad + reading to the cap & 2.70 & 0.77 & 0.72 & 0.64 & 0.67 & 77 & 99\% & 0 \\
\quad + six full reviews & 3.06 & 0.76 & 0.71 & 0.63 & 0.66 & 76 & 99\% & 0 \\
\quad + highlight stage & 2.94 & 0.75 & 0.71 & 0.64 & 0.65 & 75 & 99\% & 0 \\
\quad + cap and highlight & 3.30 & 0.77 & 0.73 & 0.67 & 0.67 & 77 & 99\% & 0 \\
\quad + written reviews & 2.64 & \multicolumn{7}{l}{placement as when keeping the hours} \\
\midrule
Two tiers, staged & 3.40 & 0.74 & 0.66 & 0.57 & 0.64 & 74 / 97 & 98\% & 0 \\
Two tiers, today & 6.0 & 0.51 & 0.20 & 0.12 & 0.36 & 51 & 81\% & 3.6\% \\
\bottomrule
\end{tabular}
\end{table}

Under the staged funnel, a broad accepted list costs exclusivity, not quality. The 25\% list holds 97 of the 100 papers in the true top 10\%, and 0.7\% of it comes from the bottom half (Table~\ref{tab:design-lever}). Today's 10\% list holds 51 of those 100 papers, and 3.6\% of it still comes from the bottom half; today's 25\% list draws 10.5\% from there. But the staged 25\% list accepts 250 papers in every thousand, and a conference that wants its accepted list to mean the top tenth will not get that from it.

Going from 25\% to 10\% under the staged funnel frees about 1.1~h per submission (3.40~h to 2.34~h): at NeurIPS 2026's scale, some 32,500~h a cycle, or 16 researcher-years. The organizers can spend some of these freed hours, spend all, or keep them. Per hour, reading up to the cap buys the most: about 0.4~h lifts the hit rate at the line from 0.75 to 0.77 and at the top from 0.69 to 0.72 (Table~\ref{tab:design-lever}). More full reviews per finalist buy less: six instead of three cost 3.06~h for 0.76 at the line, against 2.70~h for 0.77 by reading to the cap. A highlight stage, in which the top 5\% of submissions get six more full reviews before spotlights and orals are chosen, costs 0.6~h and lifts the hit rate at the top to 0.71. Spending all the hours, on the cap and a highlight stage together, buys a few points: at 3.30~h, 0.77 at the line and 0.73 at the top. The readers' cap blocks more reading, and the structure has already done the main work. Some hours can also buy what the model does not count. A written review for each paper between the 10\% and 25\% accept lines, the papers a broader conference would have accepted, costs 0.3~h. It buys no placement, but it gives those authors the feedback that the paper's design withholds from papers cut at the gist stage, and it helps them resubmit. Keeping the hours returns them to research. Which use is right depends on what the conference is for.

One ranking can also carry two accept lines. Run the 25\% design, and draw a main track at the 10\% accept line with a second tier for the next 15\%. At the broad design's 3.4~h, the main track is chosen almost as well as by the design at 10\%: 0.74 at the line against 0.75, with none of it from the bottom half (Table~\ref{tab:design-lever}). The second tier keeps the broad list's participants, and the two tiers together hold 97 of the 100 top-10\% papers. Today's funnel can draw two accept lines as well, but its main track is no better placed than its own 10\% list: 0.51 at the line, against the staged main track's 0.74.

Three effects of the lever lie outside the model. An accepted list with fewer weak papers may draw more attendance and raise the conference's prestige, or it may not; only registrations and submissions over several cycles will show which. How a second tier is presented and valued, in the proceedings and in the standing it gives its papers, decides whether its authors gain from it. And a tighter rate may raise resubmissions, which would spend some of the freed hours reading the same papers again. These turn on the conference's goals and on the community's response. The lever serves those goals; it cannot set them.

\subsection{The decision record}
\label{app:design-record}

The paper's design, then, is a set of priced choices. Structure and the readers' cap settled most of them. A running funnel's own data tune two, the gate's pass rate and the merge weight. The rest are the organizers': whether to use the gate, three full reviews per finalist, the finalist margin, and the acceptance rate, which under the staged funnel trades exclusivity for hours instead of for quality. Table~\ref{tab:design-record} records each choice, what settled it, and the measurement that would revise the assumption it rests on. All but one of those measurements come from the funnel's first cycle; what an accepted list with fewer weak papers does to attendance takes several cycles to show. An organizer who reads the record knows which settings the numbers decided, which their first cycle will tune, and which were always theirs.

\begin{table}[t]
\caption{The decision record at an acceptance rate of 25\%. ``Settled by'': structure and the hours, the readers' cap, a computed comparison, the running funnel's data, or the organizers. The last column names the assumption each choice rests on and the measurement that would revise it, from the funnel's first cycle unless the row says otherwise.}
\label{tab:design-record}
\centering
\footnotesize
\setlength{\tabcolsep}{3pt}
\begin{tabular}{p{0.15\textwidth}p{0.21\textwidth}p{0.13\textwidth}p{0.15\textwidth}p{0.28\textwidth}}
\toprule
Choice & What it trades & Setting at 25\% & Settled by & Assumption; measurement that revises it \\
\midrule
Full reviews for finalists only & full-review hours on every paper against gist reading & finalists only & structure (the hours) & peers read as well as today's reviewers; each peer's agreement with the consensus \\
Gists per reader & placement against reading hours & 16 (20 sorts per paper) & the readers' cap, and a trade within it (0.4~h against 1.5 points at the line and 3 at the top) & two sets a reader a cycle; how long sorts take \\
Set size & comparisons per read against what a reader can order & ten & the readers' cap & ten gists can be sorted reliably; how long sorts take \\
Sort or score & what a read says against how long it takes & sort & computed, at equal hours & ordering is no easier than scoring; a sample of gists both sorted and scored by different readers \\
Whether to use the gate & 0.4~h, or a point at the line and four at the top, and the full reviewers' hours spent screening, against letting an AI cut papers & a gate, the reasonable choice; optional & the organizers' judgment & the conference's policy on AI review; none \\
Gate pass rate & sorts per paper against papers cut from above the accept line & 80\% if the gate reads as well as one full review & the running funnel's data: cut about one in twenty from above the accept line & gate noise; the audit sample \\
Merge weight & gist evidence against full-review evidence & 0.87, by judgments & the running funnel's data: weight by precision & gist noise; agreement across each paper's sorts \\
Full reviews per finalist & 0.6~h each against a majority on a disputed flaw & three & the organizers' judgment & what the delivery check buys; how often full reviewers remove a finalist, and how often its full reviews disagree with its gist rank \\
Finalist margin & 0.15~h per 10\% against room to remove a finalist & 20\% & the organizers' judgment & as above \\
Acceptance rate & exclusivity against breadth and the freed hours & 25\%, or two tiers & the organizers' goals & what fewer weak papers do to attendance; registrations and submissions over several cycles \\
\bottomrule
\end{tabular}
\end{table}

The exercise can be rerun for any venue, in eight steps:
\begin{enumerate}
\item state the goals, the measures and the readers' cap;
\item calibrate the model to today's process;
\item account for the hours;
\item vary each choice around a reasonable design;
\item search the searchable choices together;
\item adopt a design in the golden middle whose gain over today survives stress, not the cheapest match;
\item set the acceptance rate to the goals, and decide what the freed hours are for;
\item name the measurement that would change each trade.
\end{enumerate}
Only the second step needs data a venue may not have. Without a consistency experiment, a venue can start from the agreement between two reviews of the same paper, which it already has. Because some review noise may be shared by everyone who reads a paper (Section~\ref{sec:risks}), that agreement overstates reliability; a sample of papers reviewed twice by independent committees, as NeurIPS did, calibrates better. What the model can price, we computed; what it cannot, we have named as the organizers' to decide.

\section{Authors and the limits}
\label{app:authors}

For the limits, the model adds 30,000 authors, each with an ability and a heavy-tailed productivity of about 0.67 papers led per cycle, so that about 20,000 papers are written a cycle. A paper's quality correlates with its lead author's ability at $\rho \in \{0.3, 0.5, 0.7\}$; how strongly it does is not known, which is why all three are shown. Each paper has on average 2.5 co-authors, drawn in proportion to productivity. The pool grows 40\% a cycle, and the model assumes that newcomers have the same distribution of ability as everyone else. The conference accepts 24.5\% of the papers written, and makes 3.2\% spotlights or orals, whatever a limit does to the number submitted. Cycle 0 runs today's funnel with no limit and seeds prior acceptance; cycles 1--3 apply the limit, and the figures are for cycle 3, averaged over six runs.

Section~\ref{sec:limits} defines the three limits. ICLR's new-author rule is applied as written, greedily, with papers in random order, and a prior acceptance stands in both for having published at a major venue and for being a qualified reciprocal reviewer. An earlier version of this work computed the rule in a lead-author form, which limits only each paper's lead author; that form binds on 1.5\% of submissions against the as-written rule's 3.6\%. Table~\ref{tab:limits} and Figure~\ref{fig:limits} use the rule as written. The newest authors' share of blocked papers barely moves with $\rho$: 68--69\% under the strict rule and 74--76\% under ICLR's rule as written. Under the self-limit, each lead author rates their own papers with noise of variance 1, against a full review's 2.45; its benefit rests on authors judging their own papers better than reviewers do. In the model a self-limit binds lead authorship only, while real caps count every authorship. Table~\ref{tab:limits}, in Section~\ref{sec:limits}, gives the limits at $\rho = 0.5$. Table~\ref{tab:authors} gives all three strengths, with both forms of ICLR's rule, the staged funnel, option A, and two of the staged designs with a gate passing 60\% from Figure~\ref{fig:frontier}, one of them with a self-limit added. At every strength, the staged funnel makes far more top-0.5\% papers spotlights or orals than today's funnel under any limit, and takes far fewer accepted papers from the bottom half.

\begin{table}[h]
\caption{Limits and staged designs in the model with authors, at three strengths $\rho$ of the link between a lead author's ability and a paper's quality (computed; third cycle, six runs). Every figure is against all papers written. ``Newest'': percent of blocked papers led by authors who entered in the latest cycle. Ability gap: mean ability of authors with a prior acceptance against those without, under today's funnel with no limit.}
\label{tab:authors}
\centering
\scriptsize
\setlength{\tabcolsep}{2pt}
\resizebox{\textwidth}{!}{%
\begin{tabular}{llrrrrrrrrr}
\toprule
$\rho$ & Funnel and limit & Sub- & Hours per & Top 3\%: & Top 3\%: & Top 0.5\%: & Top quarter: & Bottom half's & Mean quality: & Newest \\
 & & mitted & paper written & submitted & accepted & spotl.\ or oral & accepted & share & blocked vs submitted & \\
\midrule
0.3 & today, no limit & 100 & 6.0 & 100 & 92 & 37 & 62 & 10.3 & -- & -- \\
 & today, strict rule & 90 & 5.4 & 90 & 84 & 37 & 60 & 11.8 & $-0.01$ vs $0.00$ & 68 \\
 & today, ICLR's new-author rule as written & 96.5 & 5.8 & 96 & 89 & 38 & 61 & 10.8 & $-0.01$ vs $0.00$ & 74 \\
 & today, ICLR's rule, lead-author form & 98.5 & 5.9 & 98 & 89 & 36 & 62 & 10.5 & $-0.01$ vs $0.00$ & 86 \\
 & today, self-limit of two & 75 & 4.5 & 95 & 90 & 43 & 62 & 10.6 & $-0.39$ vs $+0.13$ & 28 \\
 & staged, recommended (sorting) & 100 & 3.4 & 100 & 99 & 97 & 80 & 0.7 & -- & -- \\
 & staged, option A (gate 80\%, scoring) & 100 & 3.8 & 100 & 99 & 96 & 77 & 1.4 & -- & -- \\
 & staged, gate 60\%, scoring & 100 & 3.3 & 100 & 95 & 95 & 75 & 1.8 & -- & -- \\
 & staged, gate 60\%, scoring, self-limit & 75 & 2.9 & 95 & 90 & 92 & 70 & 3.0 & $-0.38$ vs $+0.13$ & 29 \\
 & \multicolumn{10}{l}{ability gap: $+0.04$ vs $-0.03$ (0.07)} \\
\midrule
0.5 & today, no limit & 100 & 6.0 & 100 & 91 & 37 & 62 & 10.4 & -- & -- \\
 & today, strict rule & 90.2 & 5.4 & 90 & 84 & 36 & 60 & 11.3 & $-0.01$ vs $0.00$ & 68 \\
 & today, ICLR's new-author rule as written & 96.4 & 5.8 & 97 & 89 & 38 & 62 & 10.7 & $0.00$ vs $0.00$ & 76 \\
 & today, ICLR's rule, lead-author form & 98.5 & 5.9 & 98 & 90 & 35 & 62 & 10.4 & $+0.05$ vs $0.00$ & 85 \\
 & today, self-limit of two & 75.6 & 4.5 & 94 & 89 & 34 & 61 & 11.1 & $-0.34$ vs $+0.11$ & 29 \\
 & staged, recommended (sorting) & 100 & 3.4 & 100 & 99 & 97 & 80 & 0.7 & -- & -- \\
 & staged, option A (gate 80\%, scoring) & 100 & 3.8 & 100 & 99 & 97 & 78 & 1.5 & -- & -- \\
 & staged, gate 60\%, scoring & 100 & 3.3 & 100 & 95 & 94 & 75 & 1.7 & -- & -- \\
 & staged, gate 60\%, scoring, self-limit & 75 & 2.9 & 93 & 88 & 91 & 69 & 3.2 & $-0.33$ vs $+0.11$ & 29 \\
 & \multicolumn{10}{l}{ability gap: $+0.05$ vs $-0.04$ (0.10)} \\
\midrule
0.7 & today, no limit & 100 & 6.0 & 100 & 93 & 40 & 63 & 10.3 & -- & -- \\
 & today, strict rule & 90 & 5.4 & 92 & 86 & 35 & 60 & 11.2 & $-0.06$ vs $+0.01$ & 69 \\
 & today, ICLR's new-author rule as written & 96.4 & 5.8 & 97 & 89 & 35 & 61 & 10.7 & $-0.03$ vs $0.00$ & 75 \\
 & today, ICLR's rule, lead-author form & 98.6 & 5.9 & 99 & 92 & 38 & 62 & 10.3 & $-0.06$ vs $0.00$ & 84 \\
 & today, self-limit of two & 76 & 4.5 & 89 & 84 & 40 & 59 & 11.7 & $-0.25$ vs $+0.08$ & 28 \\
 & staged, recommended (sorting) & 100 & 3.4 & 100 & 99 & 97 & 80 & 0.7 & -- & -- \\
 & staged, option A (gate 80\%, scoring) & 100 & 3.8 & 100 & 99 & 96 & 77 & 1.3 & -- & -- \\
 & staged, gate 60\%, scoring & 100 & 3.3 & 100 & 96 & 96 & 75 & 1.7 & -- & -- \\
 & staged, gate 60\%, scoring, self-limit & 75 & 2.9 & 88 & 84 & 88 & 67 & 3.7 & $-0.24$ vs $+0.08$ & 29 \\
 & \multicolumn{10}{l}{ability gap: $+0.07$ vs $-0.06$ (0.13)} \\
\bottomrule
\end{tabular}}
\end{table}

\clearpage
\section{Further results}
\label{app:results}

All numbers in this appendix are computed in the model. Table~\ref{tab:compare} sets the staged funnel, with and without the AI gate, beside today's funnel; Table~\ref{tab:tiers} gives the staged funnel by quality tier, with where each tier's papers are cut; and Table~\ref{tab:frontier} gives the points of Figure~\ref{fig:frontier}. Tables~\ref{tab:recon} and~\ref{tab:leniency} back two choices in the gist stage: how a paper's twenty sorts are combined, and sorting over scoring when peers carry a personal scale.

The last stage barely affects placement (Table~\ref{tab:last}). With one to five full reviews per finalist, or with the finalists set anywhere from 25\% to 40\% of submissions, placement barely changes; with two full reviewers or one instead of three, the funnel would cost 2.8 or 2.2 hours per submission and, in the model, place papers as well. Ordering the finalists by their full reviews alone is much worse at the top: a top-0.5\% paper becomes a spotlight or oral 43--79\% of the time across one to five reviews, against 96--97\% when the gist score is included. Three reviewers are kept for the delivery check, so that a majority can settle a disputed flaw. Table~\ref{tab:stress} gathers the stress tests of Section~\ref{sec:risks}, and Table~\ref{tab:polish} the presentation illustration. The gate's pass rate of 80\% is what keeps a noisy gate's losses small: a gate four times as noisy as a full review that passed only 60\% of submissions would cut 11\% of top-0.5\% papers.

\begin{table}[h]
\caption{Today's funnel, today's with spotlights and orals chosen by score, and the staged funnel with and without the AI gate (computed). Placement is the percent of each quality tier reaching the outcome. Researcher-years returned are against today's 92 at 30,709 submissions, at 2,000 hours a year.}
\label{tab:compare}
\centering
\small
\setlength{\tabcolsep}{4pt}
\begin{tabular}{lrrrr}
\toprule
 & Today & Today, spotlights and & Staged & Staged, \\
 & & orals chosen by score & funnel & without the gate \\
\midrule
Expert hours per submission & 6.0 & 6.0 & 3.4 & 3.8 \\
Researcher-years returned at 30,709 & -- & -- & 40 & 34 \\
\midrule
Top 0.5\%: spotlight or oral & 37 & 66 & 97 & 97 \\
Top 0.5\%: accepted & 98 & 98 & 99.7 & 100 \\
0.5--3\%: accepted & 90 & 90 & 99 & 100 \\
3--10\%: accepted & 74 & 74 & 95.5 & 97 \\
10--25\%: accepted & 51 & 51 & 68.5 & 69 \\
25--50\%: accepted & 26 & 26 & 18 & 17 \\
Bottom half: accepted & 5 & 5 & 0.3 & 0.3 \\
\midrule
Reproducibility: accepted & 0.505 & 0.505 & 0.726 & 0.738 \\
Reproducibility: spotlight or oral & 0.114 & 0.236 & 0.544 & 0.537 \\
Reproducibility: oral & 0.031 & 0.115 & 0.390 & 0.381 \\
\bottomrule
\end{tabular}
\end{table}

\begin{table}[h]
\caption{The staged funnel by quality tier: percent reaching each outcome or better, and percent cut at the AI gate and at the gist stage.}
\label{tab:tiers}
\centering
\small
\begin{tabular}{lrrrrr}
\toprule
Quality tier & Accepted & Spotlight & Oral & Cut at & Cut at \\
 & & or oral & & the gate & the gist stage \\
\midrule
Top 0.5\%   & 99.7 & 97 & 46  & 0.3  & 0.0  \\
0.5--3\%    & 99   & 63 & 5   & 0.9  & 0.0  \\
3--10\%     & 95.5 & 15 & 0.1 & 2.5  & 0.8  \\
10--25\%    & 68.5 & 0.8 & 0  & 5.5  & 14.8 \\
25--50\%    & 18   & 0  & 0   & 11.6 & 57.7 \\
Bottom half & 0.3  & 0  & 0   & 32.1 & 66.6 \\
\bottomrule
\end{tabular}
\end{table}

\begin{table}[h]
\caption{The hours--placement frontier. Today's funnel with $k$ full reviews uses chairs for spotlights and orals. The staged designs with a gate passing 60\% score their gists (option A's stage at other settings). ``Sorting, 10 sorts'' puts each paper in ten sorts instead of twenty.}
\label{tab:frontier}
\centering
\footnotesize
\setlength{\tabcolsep}{3pt}
\begin{tabular}{lrrrrrr}
\toprule
Design & Hours & Top 0.5\%: & 0.5--3\%: & 3--10\%: & 10--25\%: & Accepted list \\
 & & spotl.\ or oral & accepted & accepted & accepted & reproduces \\
\midrule
Today, 2 full reviews & 4 & 31 & 84 & 67 & 47 & 0.45 \\
Today, 3 full reviews & 6 & 37 & 90 & 75 & 51 & 0.505 \\
Today, 5 full reviews & 10 & 40 & 96 & 83 & 56 & 0.58 \\
Today, 8 full reviews & 16 & 43 & 99 & 90 & 61 & 0.645 \\
Today, 10 full reviews & 20 & 41 & 99.5 & 92.7 & 63 & 0.676 \\
Today, 15 full reviews & 30 & 47 & 100 & 96.5 & 68 & 0.73 \\
Gate 60\%, 5 scores + 2 full & 1.95 & 87 & 94 & 84 & 58 & 0.59 \\
Gate 60\%, 10 scores + 3 full & 3.3 & 94 & 96 & 88 & 64 & 0.656 \\
Gate 60\%, 20 scores + 3 full & 4.8 & 98 & 95 & 90 & 70 & 0.70 \\
Option A: gate 80\%, 10 scores + 3 full & 3.8 & 96 & 99 & 93.5 & 66 & 0.696 \\
Sorting, 10 sorts + 3 full & 2.6 & 91 & 99 & 91 & 62 & 0.66 \\
Recommended: sorting, 20 sorts + 3 full & 3.4 & 97 & 99 & 95.5 & 68.5 & 0.726 \\
Recommended, without the gate & 3.8 & 97 & 100 & 97 & 69 & 0.738 \\
\bottomrule
\end{tabular}
\end{table}

\begin{table}[h]
\caption{Three ways of combining a paper's twenty sorts into one gist score (recommended design otherwise).}
\label{tab:recon}
\centering
\small
\begin{tabular}{lrrrrr}
\toprule
Combination & Top 0.5\%: & 3--10\%: & 10--25\%: & Accepted list & Spotlights and orals \\
 & spotlight or oral & accepted & accepted & reproduces & reproduce \\
\midrule
Mean rank & 96.3 & 95.4 & 68.8 & 0.727 & 0.545 \\
Mean normal score & 97.0 & 95.4 & 68.7 & 0.728 & 0.547 \\
Bradley--Terry & 97.7 & 95.7 & 69.2 & 0.732 & 0.547 \\
\bottomrule
\end{tabular}
\end{table}

\begin{table}[h]
\caption{Scoring and sorting under leniency $L$, a reader's personal offset on every score, in units of quality's variance (not measured). Accepted-list reproducibility, and in parentheses the 3--10\% tier's acceptance. Sorting is unaffected by construction; its row is the recommended design. Scoring catches up only with twice the scores, at 70\% more hours, and with leniency below about one quality variance.}
\label{tab:leniency}
\centering
\small
\setlength{\tabcolsep}{4pt}
\begin{tabular}{lrrrrr}
\toprule
Gist stage & Hours & $L = 0$ & $L = 0.5$ & $L = 1$ & $L = 2$ \\
\midrule
Sorting, 20 sorts per paper & 3.4 & 0.73 (95.5) & 0.73 (95.5) & 0.73 (95.5) & 0.73 (95.5) \\
Scoring, 10 scores per paper & 3.8 & 0.695 (93.7) & 0.679 (92.0) & 0.661 (90.9) & 0.634 (88.2) \\
Scoring, 20 scores per paper & 5.8 & 0.755 (96.5) & 0.740 (96.0) & 0.729 (95.5) & 0.703 (94.2) \\
\bottomrule
\end{tabular}
\end{table}

\begin{table}[h]
\caption{Presentation, with an effect of variance $P$ as a share of quality's variance, seen alike by every reader of the full paper (not measured; an illustration). Placement is scored against quality.}
\label{tab:polish}
\centering
\small
\setlength{\tabcolsep}{3.5pt}
\begin{tabular}{lrrrrrrr}
\toprule
 & \multicolumn{5}{c}{Today's funnel} & \multicolumn{2}{c}{Staged funnel} \\
\cmidrule(lr){2-6}\cmidrule(lr){7-8}
$P$ & Top 0.5\%: & 0.5--3\%: & 3--10\%: & Bottom half: & Accepted list & 3--10\%: & Accepted list \\
 & spotlight or oral & accepted & accepted & accepted & reproduces & accepted & reproduces \\
\midrule
0    & 36.5 & 90 & 75 & 4.9 & 0.504 & 95.4 & 0.725 \\
0.25 & 34   & 86 & 70 & 6.2 & 0.538 & 95.1 & 0.725 \\
0.5  & 32   & 82 & 66 & 7.4 & 0.564 & 94.6 & 0.725 \\
\bottomrule
\end{tabular}
\end{table}

\begin{table}[h]
\caption{The last stage, with finalists at 30\% of submissions: the number of full reviews per finalist, and what orders the finalists (the gist score and full reviews combined, the gist score alone, or the full reviews alone). With finalists at 25\% and 40\% and the combined order, the 3--10\% tier is accepted 94.6--96.1\% of the time across one to five full reviews.}
\label{tab:last}
\centering
\small
\setlength{\tabcolsep}{4pt}
\begin{tabular}{rrlrrr}
\toprule
Full reviews & Hours & Order by & Top 0.5\%: & 3--10\%: & Spotlights and \\
 & & & spotlight or oral & accepted & orals reproduce \\
\midrule
1 & 2.2 & combined & 96.3 & 95.0 & 0.519 \\
  &     & gist     & 95.4 & 94.8 & 0.519 \\
  &     & full     & 43.1 & 86.7 & 0.152 \\
2 & 2.8 & combined & 96.7 & 95.4 & 0.527 \\
  &     & gist     & 96.2 & 94.8 & 0.521 \\
  &     & full     & 58.6 & 89.2 & 0.213 \\
3 & 3.4 & combined & 96.9 & 95.2 & 0.539 \\
  &     & gist     & 95.6 & 94.8 & 0.511 \\
  &     & full     & 67.4 & 91.3 & 0.269 \\
5 & 4.6 & combined & 97.2 & 95.8 & 0.555 \\
  &     & gist     & 95.2 & 94.6 & 0.519 \\
  &     & full     & 79.2 & 93.6 & 0.340 \\
\bottomrule
\end{tabular}
\end{table}

\begin{table}[h]
\caption{The staged funnel under stress. Peer noise multiplies the noise of every peer judgment, gist sorts and full reviews alike; gist noise only that of judgments within sorts; gate noise that of the AI review. All multiples are of today's noise variance.}
\label{tab:stress}
\centering
\small
\begin{tabular}{lrrrr}
\toprule
Condition & Top 0.5\%: & Top 0.5\%: & 3--10\%: & Accepted list \\
 & accepted & spotlight or oral & accepted & reproduces \\
\midrule
Staged funnel & 99.7 & 97 & 95.5 & 0.726 \\
Peer noise 2$\times$ & 99.8 & 88 & 90.0 & 0.649 \\
Peer noise 3$\times$ & 99.8 & 82 & 85.0 & 0.597 \\
Peer noise 4$\times$ & 99.0 & 76 & 80.2 & 0.557 \\
Peer noise 5$\times$ & 98.9 & 68 & 76.9 & 0.528 \\
Peer noise 6$\times$ & 98.2 & 63 & 74.2 & 0.505 \\
Gist noise 2$\times$ & 99.9 & 89 & 90.5 & 0.654 \\
Gist noise 3$\times$ & 99.8 & 83.5 & 85.8 & 0.604 \\
Gate noise 2$\times$ & 98.7 & 96.5 & 92.6 & 0.708 \\
Gate noise 4$\times$ & 96.4 & 94 & 90.0 & 0.686 \\
Gate noise 8$\times$ & 93.4 & 90 & 87.6 & 0.668 \\
Shared noise & 99.7 & 76 & 81.5 & 0.565 \\
\midrule
Today's funnel & 98 & 37 & 74 & 0.505 \\
\bottomrule
\end{tabular}
\end{table}

\clearpage
\section{Hours at scale}
\label{app:hours}

\begin{table}[h]
\caption{Expert hours per cycle at the 30,709 submissions of NeurIPS 2026 \citep{riken2026}, at 2 hours per full review. Researcher-years are at 2,000 hours a year. The sorting-time rows vary the six minutes a gist of the recommended design.}
\label{tab:hours}
\centering
\small
\begin{tabular}{lrrrrr}
\toprule
Design & Hours per & Expert hours & Researcher- & \multicolumn{2}{c}{Returned against today} \\
 & submission & per cycle & years & hours & researcher-years \\
\midrule
Today, 3 full reviews & 6.0 & 184,254 & 92 & -- & -- \\
Staged funnel (6 min a gist) & 3.4 & 104,411 & 52 & 79,843 & 40 \\
Staged, without the gate & 3.8 & 116,694 & 58 & 67,560 & 34 \\
Staged, 3 min a gist & 2.6 & 79,843 & 40 & 104,411 & 52 \\
Staged, 12 min a gist & 5.0 & 153,545 & 77 & 30,709 & 15 \\
\bottomrule
\end{tabular}
\end{table}

The recommended design's 3.4 hours per submission are 1.6 hours of sorting, twenty sorts of six minutes for each of the 80\% of submissions the gate passes, and 1.8 hours of full reviews, three of two hours for each of the 30\% that become finalists. The audit sample adds 0.02 hours per submission, about 600 hours a cycle at this scale.

Submissions to ICLR 2027 look set to exceed 60,000 \citep{nature2026caps}, and we use that round number to illustrate the hours at such a scale. If 60,000 papers reached review, today's funnel would need about 180,000 full reviews and 360,000 expert hours, or 180 researcher-years; the staged funnel would take 204,000 hours and return 156,000, or 78 researcher-years.

\section{Extended related work}
\label{app:related}

The staged funnel rests on five commitments, and each meets its own line of prior work. It chooses a review design by calibrated simulation and judges limits on who may submit against changes to how the funnel judges. It spends a little judgment on every paper and full reviews on the finalists, behind an optional AI gate. Each submission brings a reader who owes the reading it costs. And peers rank short gists, drafted by an AI and approved by the authors, by sorting them. This appendix traces each line: what it established, the setting it assumed, and what the paper takes from it. The last subsection names the ingredients that have precedent and states what we found none for.

\subsection{Pricing a review policy before adopting it}
\label{app:related-simulation}

The case for redesigning conference review starts from two NeurIPS experiments. In 2014, two committees reviewed the same 166 papers, and their mean calibrated scores correlated at 0.55 \citep{cortes2021}. Cortes and Lawrence put half of the variation in scores down to subjective judgment. They also simulated how consistent acceptance would be at other accept rates and numbers of reviewers, cautioning that the simulation oversimplifies the final decision. In 2021, 203 of 882 duplicated papers, or 23.0\%, received inconsistent recommendations, and the committees would have chosen 29 and 25 papers for orals and spotlights while agreeing on 3 \citep{beygelzimer2023}. Beygelzimer et al.\ conclude that a more selective conference would be more arbitrary.

A second line simulates whole review systems. In an agent-based model calibrated with empirical data, Kovanis et al.\ find that journals sharing reviews match or beat the conventional system with far less reviewer time \citep{kovanis2017}. Esarey finds a strong role for chance in simulated journal review, which desk rejection before review partly mitigates \citep{esarey2017}. Zhang et al.\ trace much of a conference's reviewing load to resubmitted borderline papers \citep{zhang2022}. A scoping review finds the predictions of such models rarely tested against data \citep{feliciani2019}.

The paper builds on both lines. It uses the two experiments as calibration targets (Section~\ref{sec:model}) and takes the size of reviewer leniency from Cortes and Lawrence's fitted model \citep[Table~2]{cortes2021}. Their simulation already varies the number of reviewers per paper, the lever Section~\ref{sec:today} prices, and the 2021 finding on selectivity precedes the paper's result that a tighter acceptance rate leaves more to chance under today's funnel (Section~\ref{sec:buys}). The journal models assume many journals, resubmission among them and calibration to publication data. The paper asks a narrower question: for one conference whose readers' agreement has been measured, which design places papers better per expert hour, and does its lead hold when the unmeasured judgments are noisier than assumed (Section~\ref{sec:risks})?

Limits on who may submit have precedent beyond ICLR and CVPR. From October 2026, ACL Rolling Review caps authors at 20 submissions, 5 as first author \citep{acl2026}. NSF's DEB and IOS divisions paired a limit of two proposals per PI a cycle with a preproposal stage \citep{katz2017}. Su et al.\ report that scores calibrated by ICML 2023 authors' rankings of their own submissions estimate the expected review scores better than the raw scores do \citep{su2025}, which bears on the self-limit's assumption that authors judge their own papers better than reviewers do (Appendix~\ref{app:authors}). The paper sets the adopted limits and a changed funnel side by side in one model, with authors entering over several cycles, and counts hours saved, placement and who is blocked (Section~\ref{sec:limits}).

\subsection{Readers who come with the submissions}
\label{app:related-peers}

The idea that each applicant should bring a reader predates the flood. \citet{merrifield2009} proposed it for telescope time. Each applicant would rank $m$ proposals from their own sub-field, in the order they expect the community to rank them, or have their own proposal removed. A modified Borda count merges the rankings, and referees whose lists agree with the global list move their own proposal a modest number of places up. No individual's load rises as applications grow. In a toy simulation of 100 applicants, each unable to tell apart proposals within ten places, the error falls to that intrinsic uncertainty by $m = 8$. The authors envisage a last stage in which expert referees vet the approved proposals, and suggest the method for talk slots at an over-subscribed conference. They also write that there is no objective way to claim it would do better or worse than panels.

Observatories and a funder then ran the idea. Gemini's Fast Turnaround program committed each PI or co-investigator to review up to about ten other proposals, and removed the proposal of anyone who missed the deadline \citep{mason2014}. NSF's Sensors and Sensing Systems program ran the Merrifield--Saari mechanism in 2013: each PI ranked seven other proposals, and agreement with the global ranking could move a proposal up by at most two positions \citep{nsf2013}. At ESO, 172 volunteers each entered one proposal and graded others in a shadow run that did not affect allocation; first-quartile agreement was about 40\% for the distributed process and the panels alike \citep{kerzendorf2020}.

ALMA runs the largest system \citep{alma2026}. Each PI designates a reviewer from the proposing team, usually the PI, who ranks ten proposals assigned by an expertise-matching algorithm. A reviewer who does not finish has their own proposal rejected. Carpenter and Corvill\'on compare distributed review in Cycles 8 to 12 (8,086 proposals and 80,814 reviews) with panel review in earlier cycles, within a study of more than 20,000 proposals. The median RMS dispersion of ranks was 2.63 under distributed review and 2.59 in a simulation built from panel reviewers' independent scores before discussion; panels after discussion reached 2.40. The authors call their findings descriptive rather than causal, and note that distributed review arrived together with dual-anonymous review.

These systems established that proposers can rank one another at scale, that a duty enforced against the reviewer's own proposal secures the reading, and that the load per person stays flat as submissions grow. The staged funnel inherits all three, and takes the set of ten from ALMA. What differs is what is read and what the ranking decides. In each system the reader reads the full proposal, and the proposers' ranking or grade is the decision. In the staged funnel, peers sort gists, the gist score selects finalists, and three full reviews of each finalist check its claims and join its ranking (Section~\ref{sec:design}). The funnel also draws its sets at random, which Section~\ref{sec:risks} counts among its obstacles to collusion.

Machine-learning conferences have moved toward the same duty. Since 2025, ICLR has required every submission to have at least one author registered to review, on pain of desk rejection \citep{iclr2025}. That author reviews at least three papers, or six if they are on three or more submissions, so one person can cover several submissions. A submission is exempt when no author is qualified by an accepted paper at ICLR, NeurIPS, ICML or an equivalent venue. ICLR reports that about 20\% of its 2026 submissions had no reciprocal reviewer, and for 2027 it limits each author to one paper on which no author is a qualified reciprocal reviewer \citep{iclr2027}. NeurIPS 2025 asked every submission to nominate a reciprocal reviewer from its authors, with sanctions for late or persistently problematic reviewing that fall on the reviewer's own papers \citep{neurips2025rri}. ACL's Sustainable Reviewing Policy asks each submission to name a qualified service contributor, as reviewer or chair, who may be named on at most two submissions \citep{acl2026}. A contributor with a PhD needs at least two publications at major ACL or machine-learning venues, and submissions without a contributor enter a lottery. ACL reports that 38\% of the roughly 17,000 submissions of May 2026 came from authors with no available publication record.

These policies tie a reviewer to each submission, and the peer rule builds on them. Two conditions differ. The policies set a qualification bar, so a submission whose authors have no record brings no reviewer or enters a lottery; the peer role needs no invitation, seniority or past acceptance (Section~\ref{sec:design}). A past-publication test is the kind of credential whose incidence Section~\ref{sec:limits} measures for ICLR's new-author rule, though the paper does not simulate ACL's policy. And the policies fix the reading per person, as a number of full reviews. The peer owes the reading its own submission costs, about 1.6 gist sorts and 0.9 full reviews a cycle, and an author on several submissions reads for each unless a co-author takes the turn. The duty comes to 3.4 expert hours per submission, against today's 6, because full reviews go to the finalists (Table~\ref{tab:volumes}).

Whether unselected readers read well enough remains open. Stelmakh et al.\ report that reviewers recruited from outside the usual pool, and guided, wrote reviews that meta-reviewers rated above the conventional pool's \citep{stelmakh2021}. The funnel's peers are not selected, and the paper tests this risk by stress, not measurement (Section~\ref{sec:risks}).

\subsection{Sorting short texts instead of scoring them}
\label{app:related-sorting}

Ordinal judgment has a long record as a remedy for reviewers' private scales. Shah's survey reports that NeurIPS 2016 asked reviewers to rate 3 or better the top 30\% of papers, yet nearly 60\% of reviews did so; it lists rankings among the approaches and calls miscalibration a wide-open problem \citep{shah2022}. In peer grading, \citet{raman2014} extended Mallows, Bradley--Terry, Thurstone and Plackett--Luce models with grader reliability; in a course of about 170 students, peer grades of reports were roughly as consistent with TA grades as other TA grades were. The paper's gist score draws on the same family: it recommends Bradley--Terry \citep{bradley1952}, fitted by a minorization--maximization iteration \citep{hunter2004}, with the mean rank, which its figures use, as a simple fallback; the two place papers the same within simulation error (Appendix~\ref{app:model}).

Rankings are not more informative than scores as such. \citet{wang2019} design estimators that use scores and, even under arbitrary or adversarial miscalibration, strictly and uniformly outperform all estimators that use the ranking alone. The paper's choice is consistent with this, because it is made at equal hours. A sort of ten gists takes about an hour and a score of one gist about 15 minutes, so the hours that buy a paper twenty sorts buy it eight scores. At equal hours sorting reaches 0.81 at the line against 0.76 for scoring; at equal reads scoring does slightly better, 0.83, but costs 5.8 hours per submission against 3.4 (Appendix~\ref{app:design-reading}). Sorting's advantage is mostly time.

What peers rank is the second difference. The rankings above are of full papers, assignments or proposals. The closest precedent for a short summary in a fixed template is ASPLOS 2021's author-written abstracts, discussed below \citep{kozyrakis2021}. We found no verified work in which peers rank summaries that an AI drafts in a fixed template and the authors correct and approve. The paper's gist states the problem, the idea, the claimed result and the evidence offered for it, and leaves the evidence itself to the finalists' full reviews (Section~\ref{sec:design}). How well a peer places a gist within a sort has not been measured. The main runs give it the noise of one full review; with gist sorts three times as noisy, the funnel still accepts 85.8\% of 3--10\% papers, against 74\% today (Section~\ref{sec:risks}).

\subsection{Full reads for the finalists}
\label{app:related-staging}

Moving review effort away from likely rejections has more than a decade of practice. \citet{leytonbrown2024} trace it from ICML 2009 through OOPSLA, ASPLOS and PLDI to AAAI, IJCAI and ICML in 2022. At AAAI-21, each of 7,133 full papers first received two reviews. The 2,615 papers (37\%) with two high-confidence reviews recommending rejection were rejected; the rest went on to at least four reviews. A natural experiment put the probability that a Phase~1 reject would eventually have been accepted at 2.9\%. The authors note that Phase~1 rejects had no opportunity to rebut. They also record that NeurIPS 2020 summarily rejected 11\% of submissions after area chairs skimmed them, and dropped the practice in 2021 after negative feedback. Kim et al.'s ``two-stage'' review stages the disclosure of reviews to authors, not selection \citep{kim2025}.

ASPLOS 2021 made the first stage cheaper by changing what is read \citep{kozyrakis2021}. Authors wrote a two-page abstract answering six questions inspired by the Heilmeier Catechism: the problem, the shortcomings of prior work, the key insights, the artifacts, the results and the fit. Each abstract received at least five program-committee reviews and one external review, and about 55\% advanced to full-paper review. The chairs list open issues: the parameters need more experimentation, a flawed paper with a strong abstract can advance, and early data showing no disadvantage for non-native English speakers need a multi-year study.

Funders have run the same funnel. In 2012, NSF's DEB and IOS divisions asked for a four-page preliminary proposal, and panels invited the most meritorious to submit a full one \citep{katz2017}. About 75\% of preliminary proposals went no further, and full-proposal funding rates doubled, from 15--17\% to 28--34\%. Reviewer workload fell by 26\% in DEB and 18\% in IOS, against 24\% in the comparison programs, and the evaluators found mixed evidence that the change had furthered its goals.

A different answer to noise near the line is to stop paying for precision there. \citet{fang2016} propose that peer review identify the meritorious proposals and a lottery choose among them. Heyard et al.\ describe a Bayesian ranking at the Swiss National Science Foundation that sends proposals whose credible intervals overlap a provisional funding line to a lottery \citep{heyard2022}.

These funnels established that a cheap first stage can turn away many papers at little loss, and that a fixed template lets readers compare like with like. Merrifield and Saari's expert vetting of the winners anticipates the paper's delivery check \citep{merrifield2009}. Three conditions differ in the staged funnel. At AAAI the first stage is two full reviews, and at ASPLOS and NSF a committee's reading of an author-written summary; in the staged funnel it is twenty gist sorts by peers that the submissions bring. The first stage ranks as well as screens: 30\% of submissions become finalists, and the gist score carries into their final ranking. Ordered by their full reviews alone, the finalists would make a top-0.5\% paper a spotlight or oral 67\% of the time, against 97\% with the gist score (Section~\ref{sec:design}). And where a lottery treats the line as noise, the funnel spends cheap judgments to rank near it. Its price to authors is larger than AAAI's: a Phase~1 reject there receives two full reviews, while a paper cut before the finalists receives the gate's report, and a rank band if it passed the gate, but no written review (Section~\ref{sec:design}).

\subsection{An automated check before people read}
\label{app:related-gate}

AI reading has entered review at scale in two forms. At AAAI-26, every main-track submission, 22,977 full-review papers, received one clearly identified AI review beside at least two human reviews in Phase~1 \citep{biswas2026}. Senior program-committee members and area chairs could see it when deciding promotion to Phase~2, but it gave no scores or recommendations and replaced no human reviewer. Of 5,834 survey responses, 53.9\% judged the AI reviews useful and 20.2\% did not; the authors report errors in reading some equations and tables. Agents4Science, a venue for papers with AI agents as primary authors, let AI scores decide advancement \citep{bianchi2025}. Three LLM reviewers scored 253 complete submissions from 1 to 6, and the 79 averaging 4.0 or more went to human experts, who did not see the LLM reviews and accepted 48. Earlier studies placed AI feedback beside human review, on papers \citep{liang2024} and on reviews at ICLR 2025 \citep{thakkar2025}.

The paper's gate sits between AAAI-26 and Agents4Science. It makes a decision, turning back papers before people read them, but leaves the ranking of ideas to people. It checks form and basic soundness, whether a paper is clearly written and free of basic errors of reasoning, and does not judge the idea. Its pass rate of 80\% is a setting that an audit sample tunes, not a measured rejection rate (Section~\ref{sec:design}). The gate is optional, because some conferences will not let an AI reject papers (Appendix~\ref{app:design-trades}). Its main purpose, screening papers that could rank well on a gist without being ready for review, is one the model cannot show. Its accuracy is assumed equal to one full review's and stress-tested (Section~\ref{sec:risks}).

\subsection{Incentives and consensus: questions the paper inherits}
\label{app:related-incentives}

When the selectors are also candidates, each has a reason to rank competitors low. \citet{holzman2013} call a nomination rule impartial when no agent's message affects whether that agent wins. Partition mechanisms split the agents into groups and select within each group by the reports of agents outside it \citep{alon2011,aziz2016,xu2019}. Other impartial mechanisms avoid an explicit partition. Kurokawa et al., motivated by the NSF pilot, select a set nearly as highly rated as a non-impartial mechanism would \citep{kurokawa2015}, and PeerNomination relaxes exactness for accuracy \citep{mattei2020}. The paper names such mechanisms as a defence and leaves gaming outside its model (Section~\ref{sec:risks}).

Distributed review raises a second question that impartiality does not answer: whether it favours safe work. Merrifield and Saari name as their greatest potential concern that the process may drive allocation toward conservative mediocrity \citep{merrifield2009}. Naghizadeh and Liu's analysis of the NSF pilot argues that its process favours non-controversial proposals, and that even a coalition of two PIs can cause significant harm \citep{naghizadeh2013}. In an agent-based model, Squazzoni and Gandelli find that reciprocity between authors and referees does not by itself improve review, unless referees hold disinterested standards of fairness \citep{squazzoni2013}.

The staged funnel uses agreement with the consensus as a measurement, of how well peers judge gists and of possible collusion (Sections~\ref{sec:risks} and~\ref{sec:conclusion}). Its design includes no bonus for agreement, so Naghizadeh and Liu's analysis concerns an incentive the funnel does not use. A milder form of both concerns stays open: twenty sorts of short gists may favour work that is easy to summarize and hard to dispute. The model cannot settle this. In it, a reader's error is either drawn afresh for each judgment or shared by all readers of a paper, so no paper divides readers as a controversial proposal does (Appendix~\ref{app:model}). A taste that every reader shares counts as quality (Section~\ref{sec:model}), and the presentation runs assume that the gist's template strips presentation (Section~\ref{sec:risks}). Whether the funnel drifts toward conservative work is a question for a running funnel.

\subsection{What is and is not new}
\label{app:related-new}

Most ingredients of the staged funnel have precedent. Proposers who rank one another, with the duty enforced against their own proposal and a load per person that stays flat as submissions grow, come from \citet{merrifield2009}, Gemini \citep{mason2014}, the NSF pilot \citep{nsf2013}, ESO \citep{kerzendorf2020} and ALMA \citep{alma2026}. The set of ten comes from ALMA, and expert vetting of the winners from Merrifield and Saari. Staged, summary-first review comes from two-phase review at ICML 2009 and AAAI-21 \citep{leytonbrown2024}, ASPLOS 2021's template abstracts \citep{kozyrakis2021} and NSF's preproposals \citep{katz2017}. A reviewer tied to each submission is already required at ICLR \citep{iclr2025,iclr2027}, NeurIPS \citep{neurips2025rri} and ACL Rolling Review \citep{acl2026}. AI review of every submission ran at AAAI-26 \citep{biswas2026}, and AI scores chose which papers humans reviewed at Agents4Science \citep{bianchi2025}. Ranking in place of scoring, with Bradley--Terry-type aggregation, comes from peer grading and the study of miscalibration \citep{raman2014,shah2022,hunter2004}; scores can carry more information than rankings \citep{wang2019}, and the paper's case for sorting is made at equal hours. Calibrating to the NeurIPS experiments and simulating review systems both have precedent \citep{cortes2021,kovanis2017}.

We found no verified work that combines these parts as the staged funnel does. In it, every submission names a peer, with no qualification bar, who owes the reading that submission costs; peers sort gists that an AI drafts in a fixed template and the authors approve; and the gist ranking is merged with full reviews of the finalists, behind an optional AI gate on form and basic soundness. The design was chosen by simulation in a model calibrated to the two NeurIPS consistency experiments, priced in expert hours and set beside the submission caps and past-acceptance rules that conferences are adopting. Placement is measured against readers' shared judgment, not correctness, and the noise of the gist sorts, the unselected peers and the gate is assumed, not measured; Section~\ref{sec:risks} tests how far each can err.

\end{document}